\documentclass[11pt]{article}

\usepackage[final]{acl}
 
\usepackage{times}
\usepackage{latexsym}
\usepackage{amsmath}
\usepackage{algorithm}
\usepackage{algorithmic}
\usepackage{amssymb}  
\usepackage{booktabs}
\usepackage{multirow}
\usepackage{xcolor}
\usepackage{graphicx}
\usepackage{cuted}
\usepackage{capt-of}
\usepackage{fancyvrb}
\usepackage[T1]{fontenc}

\usepackage[utf8]{inputenc}

\usepackage{microtype}

\usepackage{inconsolata}

\usepackage{graphicx}
\usepackage{listings}
\usepackage{xcolor}
\usepackage{dsfont}
\lstdefinestyle{promptstyle}{
    basicstyle=\ttfamily\scriptsize,
    breaklines=true,
    breakatwhitespace=false,
    columns=fullflexible,
    keepspaces=true,
    frame=single,
    xleftmargin=0pt,
    xrightmargin=0pt,
    aboveskip=4pt,
    belowskip=4pt
}

\title{MangaFlow: An End-to-End Agentic Framework for \\
Controllable Story to Manga Generation}

\author{
  Muyao Wang \\
  The University of Tokyo \\
  \scriptsize\texttt{muyaowang@g.ecc.u-tokyo.ac.jp}
  \And
    Zeke Xie\thanks{Corresponding author.} \\
    Hong Kong University of Science and Technology \\
    (Guangzhou) \\
    \scriptsize\texttt{zekexie@hkust-gz.edu.cn}
  \AND
Yanhao Chen \\
The University of Tokyo \\
{\scriptsize\texttt{chen-yanhao@g.ecc.u-tokyo.ac.jp}}
\And
Lixin Xiu \\
The University of Tokyo \\
{\scriptsize\texttt{xiu-lixin106@g.ecc.u-tokyo.ac.jp}}
\And
Hideki Nakayama \\
The University of Tokyo \\
{\scriptsize\texttt{nakayama@ci.i.u-tokyo.ac.jp}}
}

\begin{document}

\maketitle

\begin{figure*}[h]
    \centering
    \begin{minipage}[t]{0.24\textwidth}
        \centering
        \includegraphics[page=1,width=\linewidth]{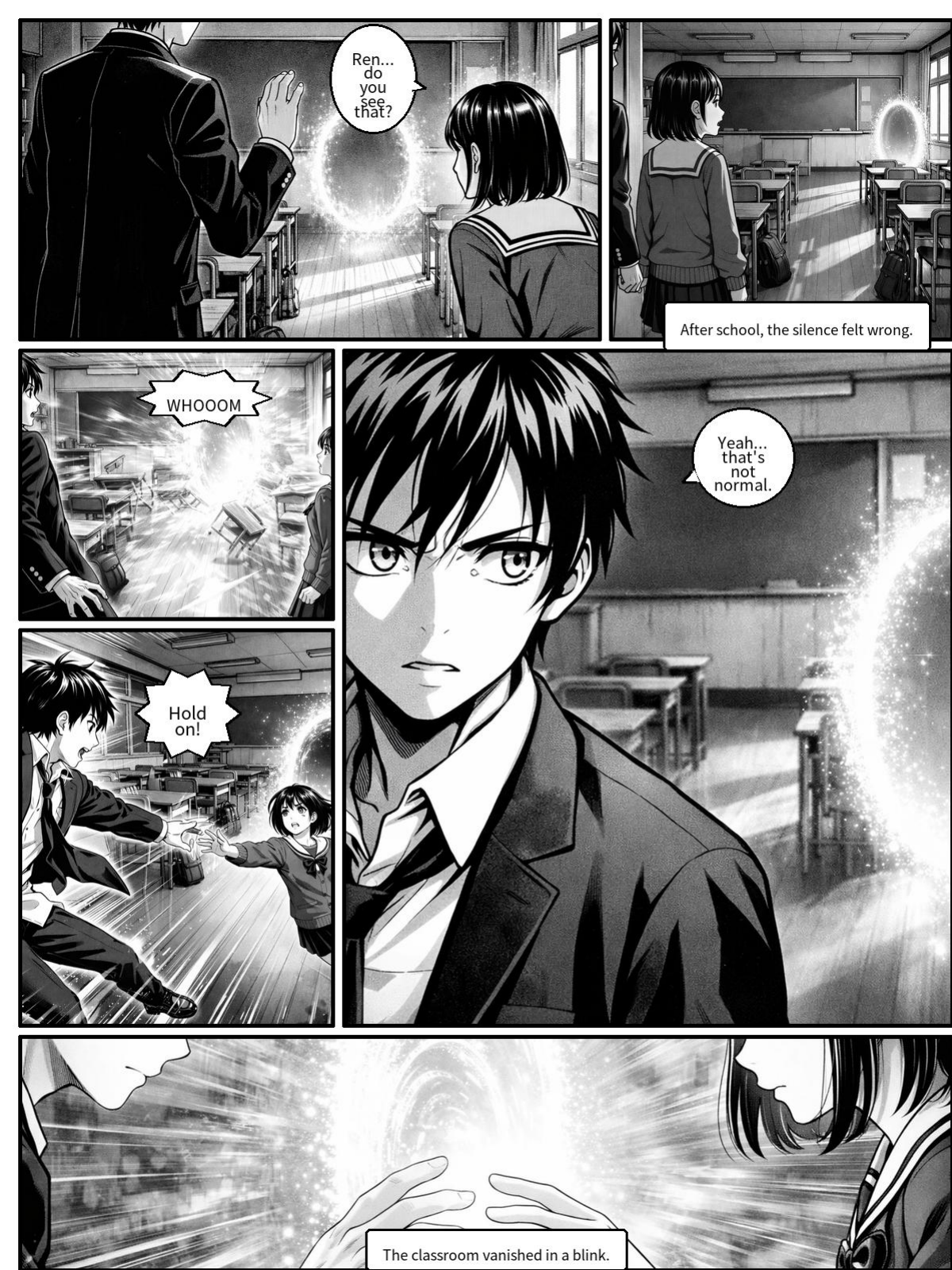}
    \end{minipage}
    \hfill
    \begin{minipage}[t]{0.24\textwidth}
        \centering
        \includegraphics[page=2,width=\linewidth]{figures/text2manga_demo.pdf}
    \end{minipage}
    \hfill
    \begin{minipage}[t]{0.24\textwidth}
        \centering
        \includegraphics[page=3,width=\linewidth]{figures/text2manga_demo.pdf}
    \end{minipage}
    \hfill
    \begin{minipage}[t]{0.24\textwidth}
        \centering
        \includegraphics[page=4,width=\linewidth]{figures/text2manga_demo.pdf}
    \end{minipage}

\caption{
Example of end-to-end text-to-manga generation by MangaFlow. 
The generated manga consists of four pages, and each page should be read from right to left following the standard manga reading order.
}
    \label{fig:text2manga_demo}
\end{figure*}

\begin{abstract}
End-to-end manga generation is a structured visual storytelling task that requires story decomposition, recurring character and scene grounding, page layout design, panel rendering, page composition, and lettering. However, existing generative models often perform direct page synthesis, entangling these factors in a single visual output and limiting precise control over layout geometry, visual references, and cross-panel consistency. To address these limitations, we propose \textbf{MangaFlow}, an agentic framework for controllable long-form manga generation that decomposes manga creation into planning, grounding, layout construction, reference-conditioned rendering, composition, and text placement. 
By treating layout and visual references as explicit intermediate variables, MangaFlow enables both simple text-to-manga generation and more precise user-controlled manga creation. This design exposes layout, visual assets, and lettering as editable intermediate controls for refining panel geometry, references, and text placement. To support long-form consistency, MangaFlow introduces a \textit{story section memory} that links section descriptions with corresponding character, scene, and object references for reuse across panels. We further present a meta-benchmark for evaluating layout controllability, visual consistency, and generation quality. Experiments show that MangaFlow improves layout adherence and cross-panel consistency over direct generation baselines while supporting flexible human control. Our code is available at \href{https://flare200020.github.io/mangaflow-page/}{MangaFlow}.
\end{abstract}

\section{Introduction}
\label{sec:introduction}

Manga generation is an emerging form of language-guided visual storytelling that goes beyond single-image synthesis. It requires a complete production process involving narrative decomposition, character and scene consistency, page-level panel layout, panel-wise rendering, and dialogue or narration placement~\cite{wu2025diffsensei,vivoli2024one,chen2024manga}. Recent story visualization methods have explored coherent image sequence generation from textual narratives~\cite{maharana2022storydall,chen2022character,zhou2024storydiffusion,rahman2023make,li2019storygan}, and manga generation methods have begun to address multi-panel page structures~\cite{chen2024manga}. However, existing approaches still provide limited support for controllable end-to-end manga production, especially for editable layouts and lettering, reusable character and scene references, and structured consistency control across panels and pages.

A central difficulty is that manga layout is not merely a visual attribute, but a core narrative device~\cite{feng2024panel,cao2012automatic,jing2015content}. Panel size, arrangement, reading order, and empty space directly affect pacing and visual emphasis. Yet direct page generation methods usually treat layout as an implicit outcome of image synthesis~\cite{xie2023boxdiff,qu2023layoutllm,inoue2023layoutdm}. As a result, they may fail to follow specified panel geometry, produce incorrect panel counts, merge panel boundaries, or ignore creator-defined layout constraints. Since artists typically design page structure before drawing panels, manga generation systems should treat layout as an explicit, editable, and enforceable structural variable.

Large Language Model (LLM)-based agents offer a natural way to decompose such a complex generation process. Prior agentic systems have shown that language models can support planning, reasoning, tool use, and multi-step coordination~\cite{yao2022react,wu2024autogen}, while visual generation agents improve controllability by decomposing complex instructions into specialized subtasks~\cite{wang2024genartist,jin2023generating,zhou2025agentstory,kim2025visagent}. Manga generation similarly requires transforming a single story prompt into structured decisions over story planning, scene segmentation, character grounding, layout construction, panel rendering, page composition, and lettering. However, generic agent pipelines are insufficient: without task-specific memory and layout control, information may decay across stages, causing character drift, scene inconsistency, object inconsistency, or layout violations.

To address these challenges, we propose \textbf{MangaFlow}, an end-to-end agentic framework for controllable manga generation. To the best of our knowledge, MangaFlow is the first agentic system designed for controllable end-to-end manga generation. MangaFlow decomposes manga creation into narrative planning, story-section construction, explicit layout generation, reference-conditioned panel rendering, page composition, and lettering. MangaFlow has two key designs: story section memory and explicit layout control. 
The former links coherent story sections with fixed textual descriptions and visual references for recurring characters, scenes, and key objects, enabling consistent reuse across panels. 
The latter treats layout as a first-class control variable, supporting manual layout specification, reference-layout reuse, and editable intermediate assets.

In addition, evaluation is another open challenge. Existing visual storytelling benchmarks mainly focus on character consistency, prompt alignment, or visual quality~\cite{zhuang2025vistorybench,gao2025vinabench}, while end-to-end manga generation also requires evaluating layout controllability, panel count correctness, lettering quality, and adherence to user-provided constraints. We therefore introduce \textbf{MangaGen-MetaBench}, a meta-benchmark covering layout-controlled generation, and story-section-guided generation. Rather than serving as a final comprehensive benchmark, it provides a practical protocol for exposing the limitations of direct generation methods and validating the controllability of MangaFlow.

Our contributions are summarized as follows:
\begin{itemize}
    \item We formulate controllable end-to-end manga generation as a structured visual storytelling task requiring narrative planning, layout control, visual consistency, and lettering.
    
    \item We propose \textbf{MangaFlow}, an agentic framework that supports explicit layout control, reference-conditioned panel rendering, and editable intermediate assets, with story section memory binding section descriptions to character, scene, and object references for cross-panel consistency.
    
    \item We propose \textbf{MangaGen-MetaBench}, a meta-benchmark for evaluating manga-specific layout controllability, consistency, generation quality, and lettering.
\end{itemize}

\section{Related Work}
\label{sec:related_work}

\paragraph{Manga Generation}
Early manga and comic generation studies mainly focus on layout design, page composition, or transforming existing visual materials into comic-style pages. 
For example, automatic manga layout and video-to-comic systems explicitly model panels, page composition, and text balloons, showing that manga creation requires more than independent image synthesis~\cite{cao2012automatic,jing2015content,yang2021automatic}. 
Large-scale manga resources and understanding benchmarks, such as Manga109~\cite{aizawa2020building}, Manga109Dialog~\cite{li2024manga109dialog}, and CoMix~\cite{vivoli2024comix}, further provide annotations for panels, characters, text, reading order, speaker association, and page-level understanding. 
More recently, MangaDiffusion~\cite{chen2024manga} formulates text-only manga generation and constructs Manga109Story to study multi-panel manga page generation from plain text, while DiffSensei~\cite{wu2025diffsensei} introduces customized manga generation with multi-character control and layout-aware character adaptation. 
These works mostly focus on isolated components such as layout conditioning, character customization, or page generation. 
MangaFlow targets the missing setting of end-to-end controllable manga creation with an editable intermediate pipeline.

\paragraph{Story Visualization}
Story visualization aims to generate a sequence of images from a textual narrative and has been studied as a natural extension of text-to-image generation. 
Early methods such as StoryGAN~\cite{li2019storygan} and StoryDALL-E~\cite{maharana2022storydall} establish the basic formulation of mapping multi-sentence stories to coherent image sequences, while later methods improve visual consistency through memory, planning, attention, or diffusion-based generation~\cite{maostory,menglogistory,zhang2025storyweaver,rahman2023make,zhou2024storydiffusion,chen2022character,shen2025storygpt,he2025dreamstory}. 
Recent benchmarks such as VinaBench~\cite{gao2025vinabench} and ViStoryBench~\cite{zhuang2025vistorybench} further evaluate story visualization from perspectives such as narrative coherence, character consistency, visual style, and prompt alignment. 
MangaFlow adapts the long-range consistency goal to manga by organizing panels into story sections.

\paragraph{Agent Systems}
LLM-based agent systems have shown strong potential for decomposing complex tasks into planning, reasoning, tool use, and multi-step execution~\cite{yao2022react,wu2024autogen}. 
In visual generation, agentic frameworks further demonstrate that specialized agents can cooperate for image generation, editing, story understanding, and narrative visualization~\cite{wang2024genartist,kim2025visagent,zhou2025agentstory}. 
Human-AI comic creation systems also show that intermediate authoring interfaces can improve creative control by allowing users to guide story structure, visual composition, or generated comic content~\cite{chen2024collaborative,jin2023generating}. 
Existing agentic visual systems mainly target general image generation/editing or story visualization, rather than manga-specific end-to-end production. 
They lack explicit control over editable page layouts, reusable visual references, and cross-panel consistency. 
MangaFlow fills this gap with an agentic framework for controllable manga generation.

\section{Methodology}
\label{sec:methodology}


\subsection{Problem Definition}
\label{sec:problem_definition}

Given a story prompt $x$, a manga generation system aims to produce a complete manga:
\begin{equation}
    \hat{M}=\{\hat{P}_1,\hat{P}_2,\ldots,\hat{P}_N\},
\end{equation}
where each page $\hat{P}_i$ contains a page layout, a set of rendered panels, and optional text elements:
\begin{equation}
    \hat{P}_i=(\hat{L}_i,\{\hat{I}_{i,j}\}_{j=1}^{n_i},\{\hat{T}_{i,j}\}_{j=1}^{m_i}).
\end{equation}
Here, $\hat{L}_i$ denotes the generated panel layout of page $i$, $\hat{I}_{i,j}$ denotes the $j$-th generated panel image, and $\hat{T}_{i,j}$ denotes a text element such as a speech bubble, narration box, thought bubble, or shout bubble.

In controllable manga generation, the system may also receive user constraints $c$, including the number of pages, target panel counts, manga style, language, user-defined layouts, reference manga layouts, character references, scene references, or key-object references. 
Formally, the generation process can be written as:
\begin{equation}
    \hat{M}=G(x,c,R),
\end{equation}
where $R$ denotes optional visual references.

\begin{figure*}[t]
    \centering
    \includegraphics[
        width=0.98\textwidth,
        trim={0 60pt 0 30pt},
        clip
    ]{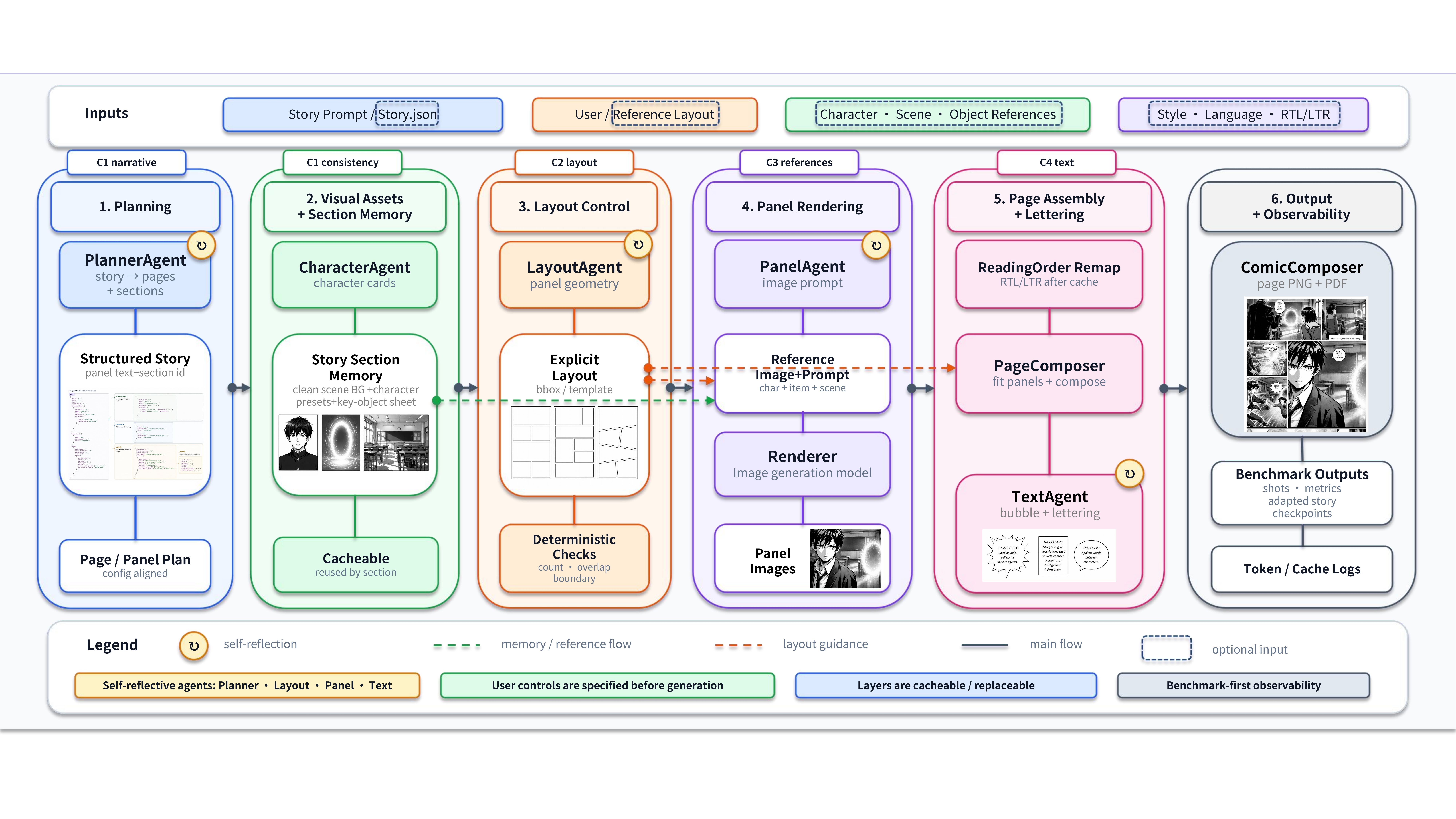}
    \caption{
    The MangaFlow framework decomposes manga generation into story planning, story-section memory construction, explicit layout control, reference-conditioned panel rendering, page composition, and lettering.
    MangaFlow can take either a simple story prompt as input or additional user-specified conditions, such as a detailed story JSON file, reference images for characters, scenes, and key objects, and target page layouts, to enable more precise and controllable manga generation.
    }
    \label{fig:framework}
\end{figure*}
\subsection{Challenges}
\label{sec:challenges}

To the best of our knowledge, MangaFlow is the first end-to-end agentic system for controllable manga generation. This setting introduces four key challenges.

\textbf{C1: Narrative decomposition and long-range consistency.} 
Manga stories involve multiple scenes, recurring characters, and key objects. 
Generating panels independently can cause character, scene, and object drift across panels and pages.

\textbf{C2: Explicit layout controllability.} 
Manga layout is a core storytelling structure rather than a purely aesthetic outcome. 
A practical system should follow user-defined panel counts and geometry while supporting reference-layout reuse.

\textbf{C3: Reference-guided panel rendering.} 
Users may provide or edit character sheets, scene references, layout, and key-object designs. 
The system should reuse these references across panels instead of relying solely on textual prompts.

\textbf{C4: Page composition and lettering.} 
A manga page is more than a set of independent images. 
Rendered panels must be composed according to target layout, while speech bubbles and narration boxes should preserve readability and narrative flow.

\subsection{Framework Overview}
\label{sec:framework_overview}

MangaFlow addresses these challenges by decomposing manga creation into a structured agentic pipeline. 
The overall framework is shown in Figure~\ref{fig:framework}. 
Given a story prompt, MangaFlow first plans the story and divides it into story sections. 
Each story section maintains its own memory, including textual descriptions and visual references for shared scenes, characters, and key objects. 
The layout agent then produces or retrieves page layouts, which are further corrected and refined. 
The panel agent constructs panel-level prompts using the story context, layout, and story-section memory. 
The renderer generates each panel image with reference conditioning. 
Finally, text agent adds text bubbles based on composed pages.

\begin{table*}[!th]
\centering
\small
\setlength{\tabcolsep}{3pt}
\caption{Design rationale for the intermediate states of MangaFlow: the control
factor each state exposes, its role in the pipeline, and the limitation of
direct page generation that motivates it.}
\label{tab:states}
\begin{tabular}{@{}p{0.155\textwidth}p{0.14\textwidth}p{0.32\textwidth}p{0.31\textwidth}@{}}
\toprule
\textbf{Intermediate state} & \textbf{Control factor} & \textbf{Role} &
\textbf{Limitation of direct page generation} \\
\midrule
Story plan & Narrative decomposition &
Assigns story events to pages and panels. &
May skip, merge, or misplace events. \\
\addlinespace[2pt]
Story Section Memory & Long-range consistency &
Reuses character, scene, and object information within coherent story sections. &
Long prompts may cause character/scene/object drift. \\
\addlinespace[2pt]
Explicit layout & Panel geometry &
Represents layouts as concrete panel regions for deterministic composition. &
May produce wrong panel counts or distorted panel boundaries. \\
\addlinespace[2pt]
Visual references & Visual grounding &
Provides reusable character, scene, and object references. &
Text-only descriptions may not preserve fine-grained identity. \\
\addlinespace[2pt]
Lettering representation & Readability &
Separates dialogue/narration placement from image rendering. &
May generate missing, unreadable, or poorly placed text. \\
\bottomrule
\end{tabular}
\end{table*}

\subsection{Story Planning and Story Section Memory}
\label{sec:story_section_memory}

To address \textbf{C1}, MangaFlow introduces \textbf{Story Section Memory}. 
Instead of treating the story as a flat sequence of panels, the planner decomposes the input prompt into pages, panels, and story sections:
\begin{equation}
    \mathcal{S}=\{s_1,s_2,\ldots,s_K\}.
\end{equation}
Each story section $s_k$ corresponds to a coherent narrative segment:
\begin{equation}
    s_k=(d_k,e_k,C_k,O_k),
\end{equation}
where $d_k$ is the textual section description, $e_k$ is the shared scene, $C_k$ is the character set, and $O_k$ is the set of key objects.

For each section, MangaFlow builds a memory:
\begin{equation}
    \mathcal{M}_k=
    (d_k,R_k^{scene},R_k^{char},R_k^{obj},\phi_k),
\end{equation}
where $R_k^{scene}$, $R_k^{char}$, and $R_k^{obj}$ denote scene, character, and object references (texts and images), and $\phi_k$ denotes structured profiles such as character presets and key-object descriptions. 
Panels in the same section reuse this memory to preserve scene, character, and object consistency.

For a panel $(i,j)$ assigned to section $z(i,j)$, the panel prompt is constructed as:
\begin{equation}
    p_{i,j}
    =
    A_{\mathrm{panel}}
    (y_{i,j},L_i,\mathcal{M}_{z(i,j)}),
\end{equation}
where $y_{i,j}$ is the panel-level story description and $L_i$ is the layout of page $i$. 
This design makes consistency an explicit reusable condition rather than an implicit property of long prompts.

\subsection{Explicit Layout Control}
\label{sec:layout_control}

To address \textbf{C2}, MangaFlow treats layout as a first-class structural variable. 
Each page layout is represented as a set of panel regions:
\begin{equation}
    L_i=\{B_{i,1},B_{i,2},\ldots,B_{i,n_i}\},
\end{equation}
$B_{i,j}$ denotes a rectangular or polygonal region.

MangaFlow supports three layout sources: manual specification, reference-layout extraction from manga pages, and layout-agent generation:
\begin{equation}
    L_i=A_{\mathrm{layout}}(q_i,n_i,c),
\end{equation}
where $q_i$ is the page-level story context and $n_i$ is the target panel number. The layout agent then performs a self-reflection step that jointly evaluates
narrative appropriateness, spatial balance, and geometric validity. As its
deterministic correction component, the layout is projected as:
\begin{equation}
    \tilde{L}_i=\Pi(L_i),
\end{equation}
where $\Pi(\cdot)$ checks panel count, overlap, page boundaries, blank regions, and layout constraints.

\subsection{Reference-Conditioned Panel Rendering}
\label{sec:panel_rendering}

To address \textbf{C3}, MangaFlow supports reference-conditioned rendering with explicit visual assets. 
Users may provide character, scene, and key-object references before generation; if absent, MangaFlow automatically constructs them through story-section memory. 
Reference manga pages can also be used as structural references by extracting their layouts, which are reused during page composition.
For each panel, MangaFlow composes the references associated with its story section:
\begin{equation}
    R_{i,j}
    =
    \mathrm{ComposeRef}
    (R_{z(i,j)}^{scene},R_{z(i,j)}^{char},R_{z(i,j)}^{obj}),
\end{equation}
where $z(i,j)$ denotes the story section assigned to panel $(i,j)$. 
The renderer generates the panel image as:
\begin{equation}
    \hat{I}_{i,j}
    =
    \mathcal{R}
    (p_{i,j},R_{i,j},w_{i,j},h_{i,j}),
\end{equation}
where $p_{i,j}$ is the panel prompt, and $w_{i,j},h_{i,j}$ are determined by the target panel region.
This design gives users explicit control over characters, scenes, objects, and layout structure.

\subsection{Page Composition and Lettering}
\label{sec:composition_lettering}

To address \textbf{C4}, MangaFlow first composes rendered panels into full pages according to the layout:
\begin{equation}
    \hat{P}_i^{img}
    =
    \mathrm{ComposePage}
    (\{\hat{I}_{i,j}\}_{j=1}^{n_i},\tilde{L}_i).
\end{equation}
The text agent then places speech bubbles, narration boxes, thought bubbles, and shout bubbles:
\begin{equation}
    \hat{P}_i
    =
    A_{\mathrm{text}}
    (\hat{P}_i^{img},\{t_{i,j}\},\tilde{L}_i).
\end{equation}
Subject centers or speaker head boxes detected by multimodal LLM guide bubble placement. 
Finally, all pages are assembled into the complete manga:
\begin{equation}
    \hat{M}
    =
    \mathrm{ComposeComic}
    (\hat{P}_1,\hat{P}_2,\ldots,\hat{P}_N).
\end{equation}

\subsection{Design Rationale for the Intermediate States}
\label{sec:rationale}

The intermediate states of MangaFlow are designed to expose manga-specific
control factors that are entangled in direct page generation. As summarized in
Table~\ref{tab:states}, each state addresses a distinct limitation of direct
generation while providing an explicit and editable representation for
downstream control. This motivates MangaFlow as a controllable
intermediate-representation pipeline rather than a direct page-generation
system.

\section{MangaGen-MetaBench}
\label{sec:benchmark}

We build \textbf{MangaGen-MetaBench} on ViStoryBench~\cite{zhuang2025vistorybench}, which provides diverse story prompts, character settings, and evaluation dimensions for narrative-driven visual generation. 
However, ViStoryBench targets image-sequence story visualization rather than manga generation, which additionally requires page-level layout control, panel structure, page composition, and lettering. 
MangaGen-MetaBench therefore augments ViStoryBench with manga-specific constraints and metrics.

\subsection{Benchmark Construction}
\label{sec:benchmark_construction}

For each story sample, we keep the original narrative prompt and character or style information from ViStoryBench when available. 
We then convert the story visualization setting into a manga generation setting by adding manga-specific page constraints. 
Each sample is associated with a page number, target panel counts and target layouts. 
The target layouts are generated by Layout agent. 
The resulting benchmark evaluates whether a method can transform a story prompt into a complete manga page sequence while respecting layout and consistency constraints.
This benchmark should be understood as a \textit{meta-benchmark} rather than a complete manga dataset. 
Its purpose is to reuse the story diversity and consistency evaluation foundation of ViStoryBench, while introducing additional manga-specific measurements.

\subsection{Metrics}
\label{sec:metrics}

MangaGen-MetaBench evaluates two groups of abilities. 
We inherit ViStoryBench metrics to measure story-level visual consistency, narrative alignment, and generation quality, and introduce manga-specific metrics for page-level layout controllability, lettering placement, and readability. 
Since the inherited metrics are not our contribution, the metric details are provided in Appendix~\ref{app:vistory_metrics}.

\paragraph{Inherited Story Visualization Metrics.}
We adopt the original ViStoryBench metrics, including \textbf{CIDS} for character identity consistency, \textbf{CSD} for style consistency, \textbf{PA} for prompt alignment, and \textbf{OCCM} for character set correctness. 
We also report \textbf{Inc}, \textbf{Aes}, and \textbf{CP} to assess image quality, aesthetic quality, and copy-paste-style pseudo-consistency. 
Together, these metrics evaluate whether generated manga panels faithfully visualize the input story and remain visually coherent.

\paragraph{Layout Controllability.}
For page $i$, let $L_i=\{B_{i,k}\}_{k=1}^{|L_i|}$ be the target layout and 
$\hat{L}_i=\{\hat{B}_{i,j}\}_{j=1}^{|\hat{L}_i|}$ be the generated or detected layout, where each $B_{i,k}$ and $\hat{B}_{i,j}$ is a panel region. 
Let $P_i$ denote the full page region and $A(\cdot)$ denote the area operator. 
We evaluate whether the generated manga follows specified page structure using following metrics.

\textbf{Panel Count Accuracy} evaluates whether the generated page contains the required number of panels:
\begin{equation}
\small
\mathrm{Acc}_{count}
=
\frac{1}{N}
\sum_{i=1}^{N}
\mathds{1}
\left[
|\hat{L}_i| = |L_i|
\right].
\end{equation}

\textbf{Layout IoU} measures the geometric agreement between the generated and target panel layouts. 
For each page, we construct a one-to-one matching $\mathcal{M}_i$ between target and generated panels by greedily selecting panel pairs in descending IoU order. 
The score is averaged over target panels, so unmatched target panels contribute zero:
\begin{equation}
\small
\mathrm{IoU}_{layout}
=
\frac{1}{N}
\sum_{i=1}^{N}
\frac{1}{|L_i|}
\sum_{(k,j)\in \mathcal{M}_i}
\frac{
A(B_{i,k} \cap \hat{B}_{i,j})
}{
A(B_{i,k} \cup \hat{B}_{i,j})
}.
\end{equation}

\textbf{Coverage Ratio} measures page-space utilization, defined as the fraction of the page covered by the union of generated panel regions:
\begin{equation}
\small
\mathrm{Coverage}
=
\frac{1}{N}
\sum_{i=1}^{N}
\frac{
A\left(
\bigcup_{j=1}^{|\hat{L}_i|}
\hat{B}_{i,j}
\right)
}{
A(P_i)
}.
\end{equation}

\textbf{Overlap Ratio} measures structural invalidity caused by overlapping panels. 
It is computed as the fraction of page area covered by two or more generated panels:
\begin{equation}
\small
\mathrm{Overlap}
=
\frac{1}{N}
\sum_{i=1}^{N}
\frac{
A\left(
\left\{
p \in P_i :
\sum_{j=1}^{|\hat{L}_i|}
\mathds{1}[p \in \hat{B}_{i,j}]
\ge 2
\right\}
\right)
}{
A(P_i)
}.
\end{equation}

\paragraph{Lettering and Readability.}
We evaluate lettering and readability with two metrics. 

\textbf{Bubble Placement Score} is evaluated by human annotators. 
For each panel containing text bubbles, annotators check whether any bubble occludes a character's face. 
A panel receives 0 if such face occlusion occurs, and 1 otherwise. 
The final score is normalized over all evaluated panels:
\begin{equation}
\small
\mathrm{BPS}
=
\frac{1}{|\mathcal{P}^{txt}|}
\sum_{p \in \mathcal{P}^{txt}}
\mathds{1}[\mathrm{no\ face\ occlusion}(p)],
\end{equation}
where $\mathcal{P}^{txt}$ denotes panels containing text bubbles.

\textbf{Manga Readability Score} is evaluated with a multimodal LLM. 
The model first reads the generated manga and summarizes the story, then compares the summary with the ground-truth story. 
The score is assigned on a 1--5 scale according to how well the generated manga communicates the original story. 
Detailed annotation and judging protocols are provided in Appendix~\ref{app:lettering_readability}.

\section{Experiments}
\label{sec:experiments}

\subsection{Experimental Setup}
\label{sec:exp_setup}

MangaGen-MetaBench extends ViStoryBench~\cite{zhuang2025vistorybench}, including 80 stories with manga-specific layout and lettering evaluations. 
The inherited metrics measure whether generated panels faithfully visualize the story, while our added metrics evaluate whether the outputs satisfy manga-specific requirements.

\paragraph{Compared Methods.}
We compare two categories of methods. 
The first category is direct page generation, where the same image generation backbones used in MangaFlow receive the full page description, including the target layout and panel information, and directly generate the whole manga page. 
This category tests whether current image generation models can follow explicit manga page specifications. 
Specifically, we use Gemini 2.5 Flash Image~\cite{google_gemini25flashimage} and FLUX.2 9B~\cite{blackforestlabs_flux2dev} as direct generation baselines. 
The second category is MangaFlow with different renderers, where the same generation backbones are used as panel renderers within our structured pipeline.

\paragraph{Metrics.}
We report three metric groups: inherited ViStoryBench metrics for story visualization quality, layout metrics for target page-structure adherence, and human evaluation for bubble placement and LLM evaluation for readability.
Details can be found in Appendix~\ref{app:vistory_metrics}

\subsection{Main Results}
\label{sec:main_results}

\begin{table}[h]
\centering
\scriptsize
\setlength{\tabcolsep}{2.5pt}
\caption{
CIDS and CSD denote self character similarity and self style similarity, respectively. 
PA is the average prompt-alignment score. 
CM denotes OCCM, Inc denotes inception score, and Aes denotes aesthetic score.
Please refer to Table.~\ref{tab:main_vistorybench} for full metric results.
}
\label{tab:main_vistorybench_compact}
\begin{tabular*}{\columnwidth}{@{\extracolsep{\fill}}lcccccc@{}}
\toprule
Method 
& CIDS$\uparrow$
& CSD$\uparrow$
& PA$\uparrow$
& CM$\uparrow$
& Inc$\uparrow$
& Aes$\uparrow$ \\
\midrule
Direct-Gem. 
& 0.562 
& 0.702 
& 2.708 
& 51.60 
& 8.14 
& 4.52 \\

Direct-FLUX 
& 0.591 
& \textbf{0.750} 
& 2.037 
& 39.05 
& 2.24 
& 4.56 \\

MF-Gem. 
& \textbf{0.643} 
& 0.622 
& 2.901 
& \textbf{73.93} 
& \textbf{12.26} 
& 5.54 \\

MF-FLUX 
& 0.619 
& 0.668 
& \textbf{3.299} 
& 66.87 
& 10.61 
& \textbf{5.92} \\
\bottomrule
\end{tabular*}
\end{table}

\begin{table}[!h]
\centering
\scriptsize
\setlength{\tabcolsep}{2pt}
\caption{
Manga-specific metrics. 
Layout metrics evaluate panel count, layout geometry, page coverage, and panel overlap. 
Bubble denotes a human-evaluated score for face-occlusion-free bubble placement, while Read. denotes a GPT-4.1-based readability score. Bubble scores are left blank for direct baselines because Direct-FLUX.2 often produces unreadable text, while Direct-Gemini frequently omits required dialogue.
}
\label{tab:main_manga_specific}
\begin{tabular*}{\columnwidth}{@{\extracolsep{\fill}}llcccccc@{}}
\toprule
\multirow{2}{*}{System} 
& \multirow{2}{*}{Renderer} 
& \multicolumn{4}{c}{Layout Metrics} 
& \multicolumn{2}{c}{Text Metrics} \\
\cmidrule(lr){3-6} \cmidrule(lr){7-8}
& 
& Count $\uparrow$ 
& IoU $\uparrow$ 
& Cov. $\uparrow$ 
& Ovl. $\downarrow$ 
& Bubble $\uparrow$ 
& Read. $\uparrow$ \\
\midrule
Direct & Gemini & 27.94$\%$ & 42.77$\%$ & 88.70$\%$ & N/A & N/A & 3.80 \\
Direct & FLUX.2 & 44.20$\%$ & 41.11$\%$ & 73.65$\%$ & N/A & N/A & 2.67 \\
\midrule
\multirow{2}{*}{MangaFlow} 
& Gemini 
& \multirow{2}{*}{\textbf{100$\%$}$^\dagger$} 
& \multirow{2}{*}{\textbf{100$\%$}$^\dagger$} 
& \multirow{2}{*}{\textbf{99.98$\%$}} 
& \multirow{2}{*}{0.62$\%$} 
& \textbf{97.4}$\%$ & \textbf{4.87} \\
& FLUX.2 
& 
& 
& 
& 
& 97.1$\%$ & 4.71 \\
\bottomrule
\end{tabular*}

\vspace{2pt}
\vspace{2pt}
\begin{minipage}{\columnwidth}
\footnotesize
\raggedright
$^\dagger$ For MangaFlow, the marked layout scores are deterministic due to explicit page composition. The two MangaFlow variants share the same explicit layouts and deterministic page composition, so their layout metrics are identical; only the rendering backbone differs.
For direct baselines, Ovl. is marked as N/A because they do not generate explicit panel layouts, making panel overlap not directly applicable. 

\end{minipage}
\end{table}

We compare direct page generation baselines with MangaFlow using different rendering backbones. 
Table~\ref{tab:main_vistorybench_compact} reports inherited ViStoryBench metrics, while Table~\ref{tab:main_manga_specific} reports manga-specific metrics for layout controllability and readability. 
On ViStoryBench, MangaFlow achieves better performance than direct generation baselines on most metrics. 
Although Direct-FLUX obtains the highest self-style consistency, its weaker performance on other metrics suggests that direct page generation may preserve local style but still struggles with story alignment and generation quality.

Table~\ref{tab:main_manga_specific} shows that explicit page composition is crucial for controllable manga generation. 
Direct baselines struggle to recover panel counts and layout geometry from prompts alone, whereas MangaFlow follows the target page structure through deterministic composition. 
MangaFlow also achieves high page coverage and supports readability evaluation, while direct baselines often fail to produce reliable dialogue or bubble structures. 
These results show that MangaFlow's structured decomposition improves layout controllability and readability without sacrificing story visualization quality.
Although the main experiments use ViStoryBench samples with detailed story and character specifications, MangaFlow also supports a simpler text-to-manga setting, where plans, layouts, references, and pages are generated automatically from a natural-language prompt. 
A qualitative example is shown in Fig.~\ref{fig:text2manga_demo}, with details in Appendix~\ref{app:text2manga_demo}. 

\paragraph{Statistical significance and human validation.}
We further validate the main results with paired story-level statistical tests
(Appendix~\ref{sec:stats}) and a blind human study of readability
(Appendix~\ref{sec:human}). The representative improvements remain significant
after multiple-comparison correction. While human annotators are stricter than
GPT-4.1 on absolute readability scores, they strongly agree with its relative
comparison between methods ($96.0\%$ no-conflict agreement), and prefer
MangaFlow in $24/25$ paired stories.

\begin{figure*}[!th]
    \centering
        \includegraphics[
        width=\textwidth,
        trim=0 50 0 0,
        clip
    ]{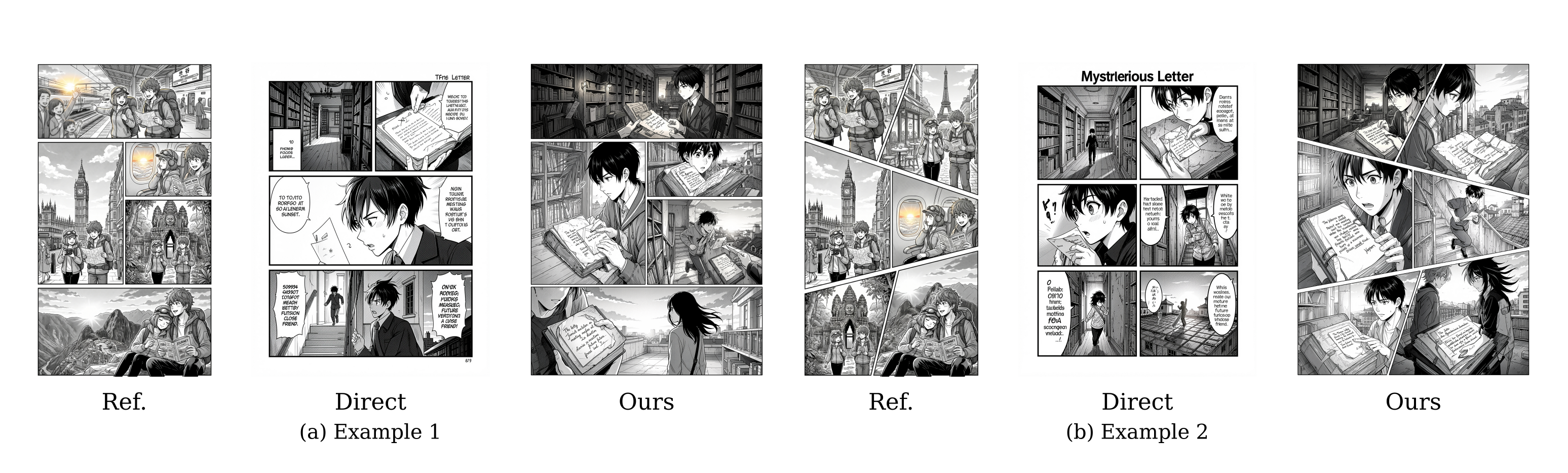}

    \caption{
    Visualization of reference-layout-guided manga generation. 
    Each example contains a reference manga layout, a direct generation result, and the MangaFlow result. 
    Direct page generation may fail to preserve the reference panel structure even when reference image is provided in the prompt, while MangaFlow explicitly reuses the extracted layout and generates new manga pages with more faithful panel arrangements.
    }
    \label{fig:reference_layout_vis}
\end{figure*}

\subsection{Ablation Study}
\label{sec:ablation}

We conduct two ablation studies to analyze the contribution of key components in MangaFlow. 
First, we evaluate \textbf{Story Section Memory}. 
This ablation is evaluated with ViStoryBench consistency and quality metrics. 
Second, we evaluate \textbf{Layout Self-Reflection}. 
Since this component directly affects page structure, we evaluate it using page-space coverage and panel overlap. 
For each ablation, we report only the metrics that are directly related to the removed component, avoiding unrelated comparisons such as evaluating story-section memory with layout IoU.

\begin{table}[h]
\centering
\scriptsize
\setlength{\tabcolsep}{3.5pt}
\caption{
Ablation study based on FLUX.2 render on Story Section Memory.
Please refer to Table.~\ref{tab:fullablation_story_section_memory} for full metric results.
CSD and CIDS are self-consistency scores.
PA denotes prompt alignment, where Scene, Shot, CI, and IA correspond to scene alignment, camera/shot alignment, character-action alignment, and individual-action alignment, respectively.
}
\label{tab:ablation_story_section_memory}
\begin{tabular*}{\columnwidth}{@{\extracolsep{\fill}}lccccccc@{}}
\toprule
Variant
& CSD$\uparrow$
& CIDS$\uparrow$
& Scene$\uparrow$
& Shot$\uparrow$
& CI$\uparrow$
& IA$\uparrow$
& Aes$\uparrow$ \\
\midrule
w/o Sec. Mem.
& 0.547
& 0.582
& 3.836
& \textbf{3.176}
& 3.439
& 2.525
& 5.83 \\

Full MF-FLUX
& \textbf{0.668}
& \textbf{0.619}
& \textbf{3.861}
& 3.008
& \textbf{3.638}
& \textbf{2.688}
& \textbf{5.92} \\
\bottomrule
\end{tabular*}
\end{table}

\paragraph{Story Section Memory.}
We remove Story Section Memory and keep the rest of the pipeline unchanged. 
Table~\ref{tab:ablation_story_section_memory} shows that the full MangaFlow-FLUX model improves both self-consistency metrics, including CSD and CIDS, compared with the variant without memory. 
It also achieves better results on most prompt-alignment dimensions, including Scene, CI, and IA, as well as a higher aesthetic score. 
This suggests that Story Section Memory helps maintain recurring characters and visual context across panels while supporting panel-level semantic alignment.

\begin{table}[h]
\centering
\small
\setlength{\tabcolsep}{6pt}
\caption{
Ablation on layout self-reflection.
We evaluate its effect using page-space coverage and panel overlap.
}
\label{tab:ablation_self_reflection}
\begin{tabular*}{\columnwidth}{@{\extracolsep{\fill}}lcc@{}}
\toprule
Variant 
& Cov. $\uparrow$ 
& Ovl. $\downarrow$ \\
\midrule
w/o Layout SR 
& 98.78\% 
& 0.86\% \\
Full MangaFlow 
& \textbf{99.98\%} 
& \textbf{0.62\%} \\
\bottomrule
\end{tabular*}
\end{table}

\paragraph{Layout Self-Reflection.}
Table~\ref{tab:ablation_self_reflection} evaluates layout self-reflection. 
Removing it lowers page coverage and increases panel overlap, showing that this refinement step improves page-space utilization and produces cleaner panel geometry. 
We therefore keep layout self-reflection in MangaFlow and provide qualitative examples in Appendix~\ref{app:self_reflection_vis}.

\begin{figure*}[h!]
    \centering

    \begin{minipage}[t]{0.48\linewidth}
        \centering
        \includegraphics[
            width=\linewidth,
            trim={0 0 0 0},
            clip
        ]{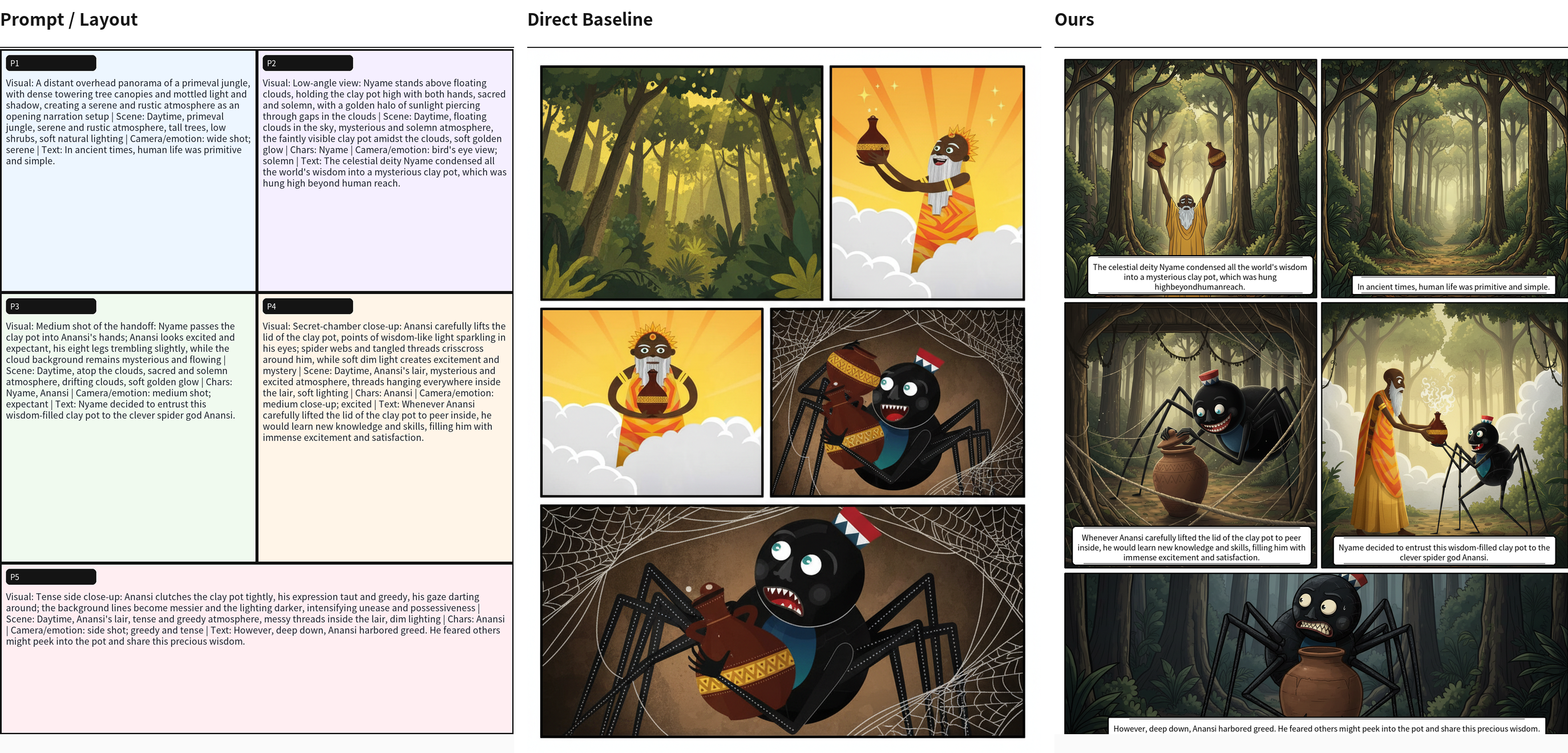}

        \vspace{2pt}
        {\scriptsize (a) Example 1}
    \end{minipage}
    \hfill
    \begin{minipage}[t]{0.48\linewidth}
        \centering
        \includegraphics[
            width=\linewidth,
            trim={0 0 0 0},
            clip
        ]{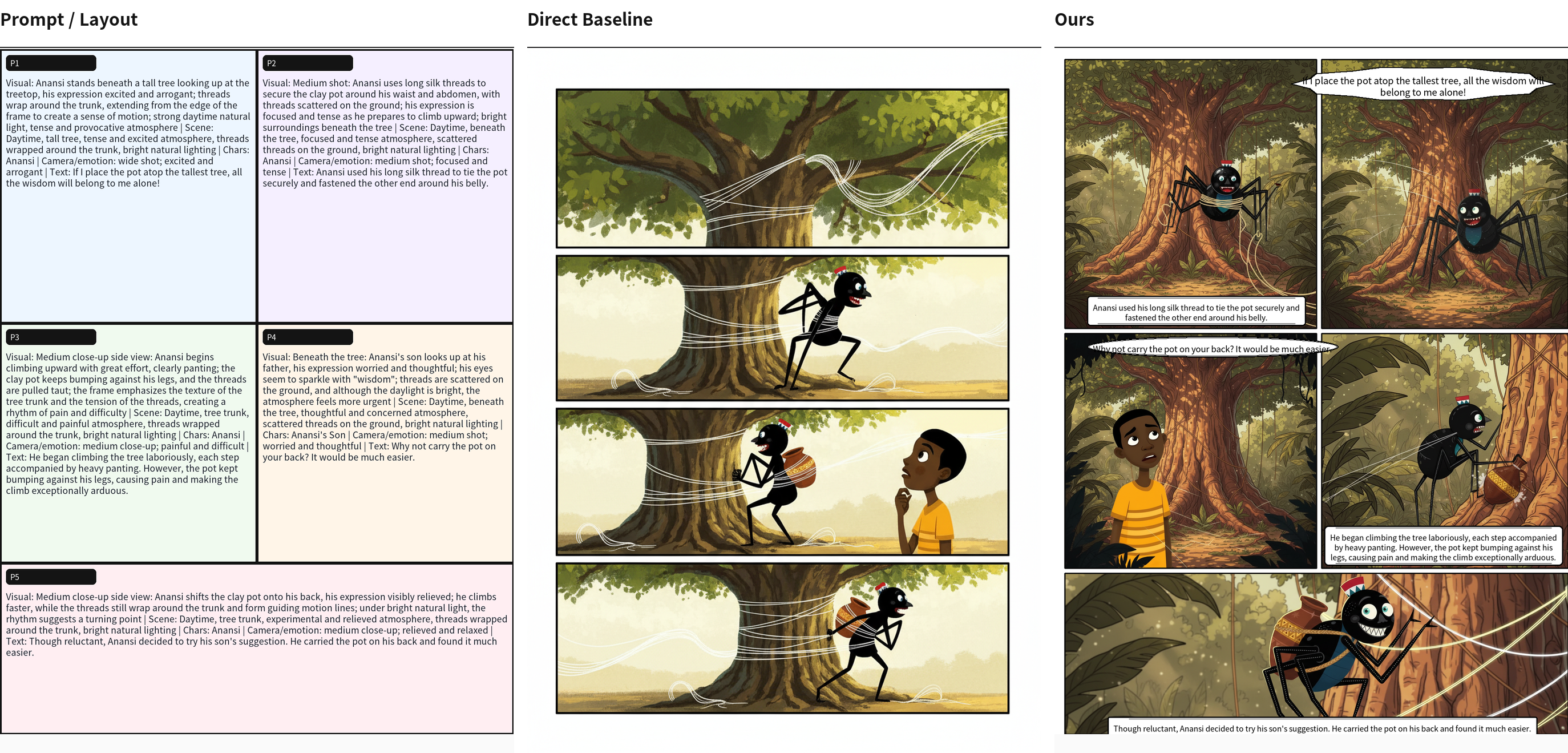}

        \vspace{2pt}
        {\scriptsize (b) Example 2}
    \end{minipage}

\caption{
Qualitative comparison of layout-controlled manga generation. 
For each example, the left side visualizes the specified target layout together with the prompt assigned to each panel, the middle shows the result of a direct page generation method, and the right side shows the result of MangaFlow. 
Although the direct method receives the same layout and panel-level prompt information, it may fail to preserve the required panel count or panel geometry, while MangaFlow follows the explicit layout through deterministic page composition.
}
    \label{fig:layout_vis}
\end{figure*}

\subsection{Visualization}
\label{sec:visualization}

\begin{figure*}[h!]
    \centering

    \begin{minipage}[t]{0.48\linewidth}
        \centering
        \includegraphics[
            width=\linewidth,
            trim={0 0 0 0},
            clip
        ]{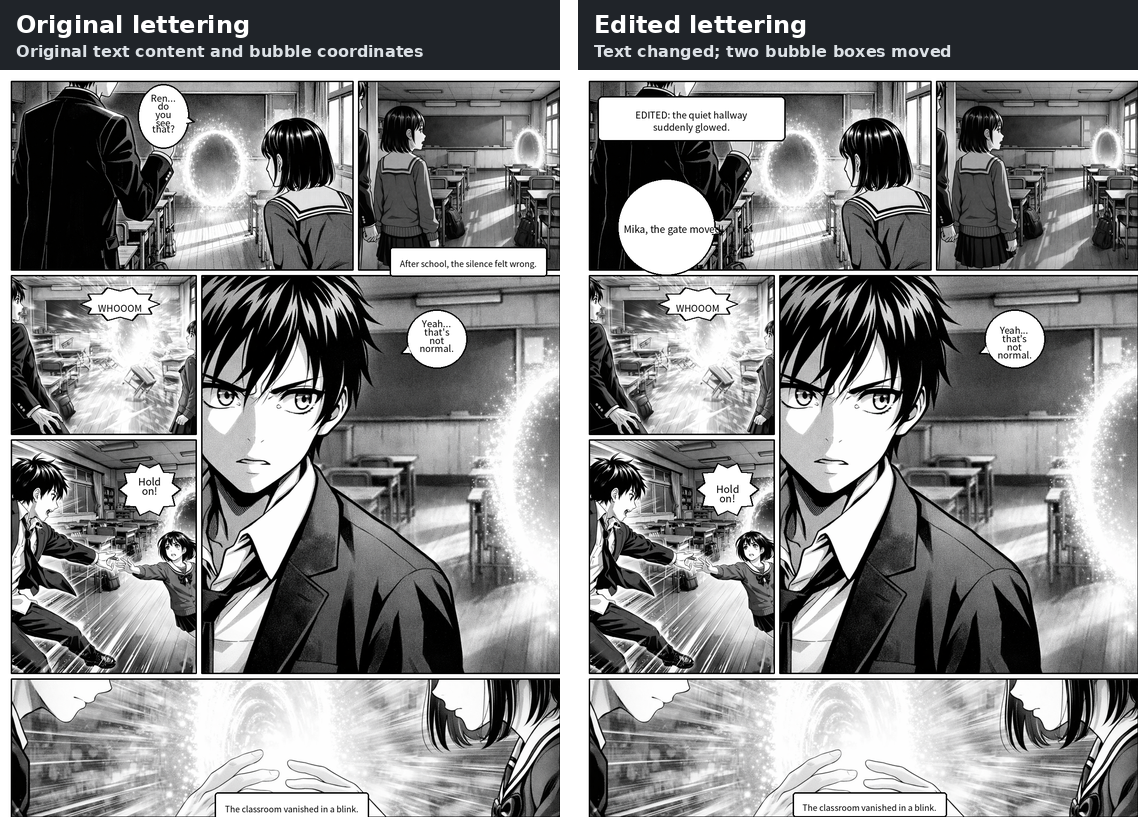}

        \vspace{2pt}
        {\scriptsize (a) Text-placement editing}
    \end{minipage}
    \hfill
    \begin{minipage}[t]{0.48\linewidth}
        \centering
        \includegraphics[
            width=\linewidth,
            trim={0 0 0 0},
            clip
        ]{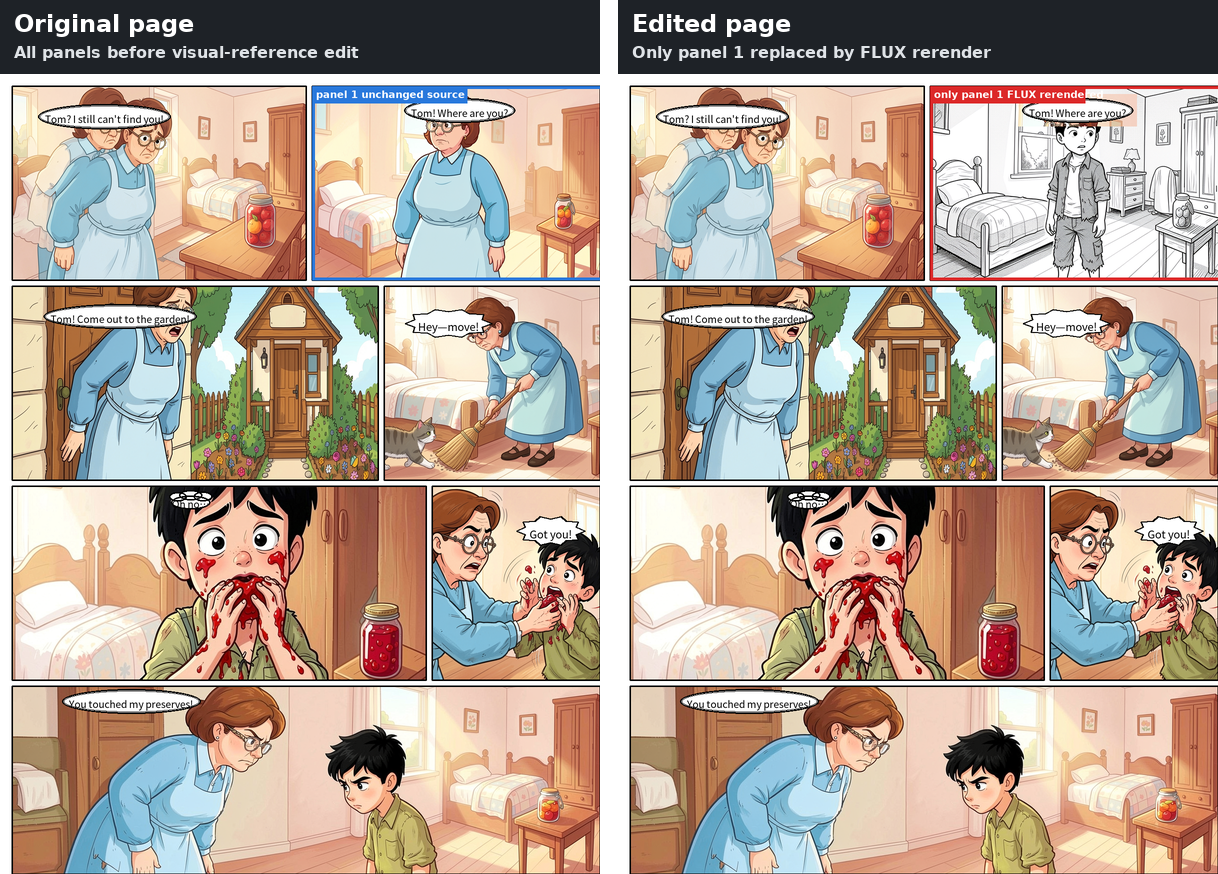}

        \vspace{2pt}
        {\scriptsize (b) Visual-reference editing}
    \end{minipage}

    \caption{
    Qualitative examples of explicit editing in MangaFlow.
    (a) Text placement and content can be directly modified through the
    lettering specification and are deterministically rendered by the
    compositor.
    (b) Visual references can be replaced explicitly, and the updated
    reference is directly propagated to the panel renderer.
    }
    \label{fig:deterministic_editing}
\end{figure*}
\paragraph{Reference-layout-guided Generation.}
We further visualize whether different methods can follow layouts extracted from reference manga. 
In Figure~\ref{fig:reference_layout_vis}, although the direct method receives the same reference manga layout information, it may fail to preserve the exact panel structure. 
In contrast, MangaFlow explicitly reuses the extracted layout as a structural constraint and generates new manga pages with more faithful panel arrangements.

\paragraph{Layout-controlled Generation.}
We visualize the difference between direct page generation and MangaFlow under the same story prompt and target layout. 
The comparison in Figure~\ref{fig:layout_vis} highlights that direct generation methods fail to preserve the required panel count or panel geometry, while MangaFlow follows the explicit layout by construction.

\paragraph{Text-placement editing.}
Text bubbles and narration boxes are placed according to explicit coordinates
and text specifications. If the bubble position or text content is edited from
$T$ to $T'$, the lettering component directly uses $T'$, and the edited bubbles
and text are rendered at the specified coordinates by the same compositor. The
edited text placement is thus deterministically executed
(Fig.~\ref{fig:deterministic_editing}(a)).

\paragraph{Visual-reference editing.}
Visual references are explicit inputs to the panel renderer. To visualize the
editing capability, we replace the reference image used by one selected panel. The corresponding panel is
re-rendered with the new reference, whereas the remaining panels are preserved
(Fig.~\ref{fig:deterministic_editing}(b)).

\section{Conclusion}
\label{sec:conclusion}

We introduced \textbf{MangaFlow}, an end-to-end agentic framework for language-guided and layout-controllable manga generation. 
MangaFlow treats manga creation as a structured generation process that coordinates story planning, section-level memory, visual references, explicit layouts, panel rendering, page composition, and lettering. This formulation provides a controllable generation paradigm for long-form manga generation. 
By making layout and references explicit intermediate variables, MangaFlow supports both simple text-to-manga generation and more precise generation with user-specified layouts or visual assets.

We further proposed \textbf{MangaGen-MetaBench}, a meta-benchmark for evaluating manga generation beyond image-sequence story visualization. 
By incorporating manga-specific dimensions such as layout adherence, page composition, lettering, and readability, MangaGen-MetaBench provides an initial evaluation protocol for controllable manga generation. 
We hope MangaFlow and MangaGen-MetaBench can serve as a step toward more structured, controllable, and evaluable systems for long-form manga generation.

\section*{Acknowledgement}
This work was supported by JSPS KAKENHI Grant Numbers JP23K28139 and JP25H01149.

\section*{Limitations}
MangaFlow demonstrates that controllable manga generation benefits from explicit intermediate structures, but several limitations remain. 
First, the final visual quality still depends on the underlying image generation backbone, so failures in character rendering, fine-grained actions, or local visual details may affect the generated panels. 
Second, although MangaFlow introduces explicit layout control and lettering, speaker-aware bubble placement remains challenging in complex panels, especially when stylized manga faces are difficult to localize. 
Third, MangaGen-MetaBench is designed as a meta-benchmark built on existing story visualization data, rather than a fully annotated manga dataset with human-designed layouts, bubble positions, and speaker annotations.

\section*{Ethical Considerations}
Manga generation systems may raise concerns about copyright, artist consent, and style imitation, especially when user-provided references resemble the work of specific artists. 
In practical use, reference images should be licensed or user-owned, and generated outputs should not be presented as human-created artwork without disclosure. 
The system may also generate inappropriate or biased visual content depending on the underlying generator and input story, so content filtering and provenance mechanisms are important for deployment.
 

\newpage
\bibliography{custom}
\clearpage
\newpage
\appendix

\section{Appendix}
\label{sec:appendix}

\subsection{Experiment and Method details}

For all external models and datasets, we follow their original terms of use and cite the corresponding creators or model providers. We use these artifacts only for research evaluation. The LLM used for our agents is GPT 5.4-nano.

Due to copyright-protection constraints of the Nano-Banana image generation API, a subset of ViStoryBench stories could not be completed with this renderer. 
As a result, all Nano-Banana-related results are computed on the 62 stories for which the API returned valid outputs. 
For a fair comparison, the corresponding direct Nano-Banana baseline and MangaFlow-Nano-Banana variant are evaluated on the same subset.

\subsubsection{Statistics of Experiment Result}
We report aggregate metrics over the full evaluation set and clarify that automatic results are from one complete generation run per method, not the maximum over multiple trials. For human evaluation, we report mean scores across annotators and samples.

\subsubsection{Computational Budget}
\label{app:compute_budget}

We evaluate $80$ stories in MangaGen-MetaBench. 
FLUX.2 9B experiments are conducted using local inference with a single NVIDIA A6000 Ada GPU. The total computational budget for FLUX.2-based generation is approximately 24 GPU hours. 

\subsubsection{Vistory Metrics}
\label{app:vistory_metrics}

To comprehensively evaluate the performance of story generation models, we adopt the evaluation framework introduced by ViStoryBench\cite{zhuang2025vistorybench}, which comprises a suite of multi-dimensional automated metrics across six key perspectives.

\textbf{Character Identification Similarity (CIDS):} This metric evaluates identity consistency by computing the cosine similarity between character features. It calculates both \textit{Cross-CIDS} (similarity between generated images and reference images) and \textit{Self-CIDS} (consistency across generated frames). The pipeline utilizes Grounding DINO for character detection, followed by feature extraction using FaceNet/ArcFace for realistic characters and CLIP ViT-L/14 for stylized characters.

\textbf{Style Similarity (CSD):} To measure stylistic coherence independent of semantic content, this metric utilizes a style-trained CLIP vision encoder combined with Content-Style Disentanglement (CSD) layers. It evaluates both \textit{Cross-CSD} and \textit{Self-CSD} via pairwise cosine similarity of the extracted style embeddings.

\textbf{Prompt Alignment:} Visual-Language Models (VLMs) are employed as automated evaluators to score the semantic alignment between generated images and text prompts on a 0-4 Likert scale. This is sub-divided into four fine-grained aspects: \textit{Scene Score} (environment and background fidelity), \textit{Shot Score} (camera perspective and composition), \textit{Character Interaction} (relational dynamics between characters), and \textit{Individual Action} (accuracy of specific gestures and poses). In our paper, we use GPT4.1 as automated evaluator. 

\textbf{Onstage Character Count Matching (OCCM):} This metric quantifies the model's ability to generate the correct number of characters as specified in the script. It applies a non-linear exponential penalty to the relative error between the detected character count $D$ and the expected count $E$:
$$OCCM = \exp\left(-\frac{|D-E|}{\epsilon+E}\right) \times 100\%$$

\textbf{Copy-Paste Detection:} To penalize models that bypass true generation by simply replicating input references, this metric calculates a Softmax-based Copy-Paste Rate. It assesses whether a generated character feature is excessively close to a specific anchor reference in the latent space, indicating overfitting rather than generalized learning.

\textbf{Aesthetic Quality and Diversity:} Finally, generation quality is assessed using the \textit{Inception Score (IS)} to measure visual diversity and clarity, along with an \textit{Aesthetic Score} evaluated via an aesthetic predictor to quantify the overall artistic appeal and perceptual quality of the generated sequence.

\begin{table}[h]
\centering
\small
\setlength{\tabcolsep}{3pt}
\caption{Compared methods. ``Direct'' means generating the whole manga page in one step. ``Pipeline'' means generating panels through MangaFlow and composing them with explicit layouts.}
\label{tab:methods}
\begin{tabular}{lccc}
\toprule
Method & Type & Layout & Sec. Mem. \\
\midrule
Direct-Gemini & Direct & Prompt & -- \\
Direct-FLUX.2 & Direct & Prompt & -- \\
MangaFlow-Gemini & Pipeline & Explicit & \checkmark \\
MangaFlow-FLUX.2 & Pipeline & Explicit & \checkmark \\
\bottomrule
\end{tabular}
\end{table}

\begin{table}[!h]
\centering
\small
\setlength{\tabcolsep}{3pt}
\caption{Metric groups in MangaGen-MetaBench. ViStoryBench metrics are inherited, while layout and lettering/readability metrics are added for manga generation.}
\label{tab:metric_groups}
\begin{tabular}{ll}
\toprule
Group & Metrics \\
\midrule
ViStoryBench & CIDS, CSD, PA, OCCM, Inc, Aes, CP \\
Layout & Count Acc., Layout IoU, Coverage, Overlap \\
Human and LLM & Bubble Placement, Readability \\
\bottomrule
\end{tabular}
\end{table}

\begin{algorithm}[h]
\small
\caption{MangaFlow Generation Pipeline}
\label{alg:mangaflow}
\begin{algorithmic}[1]
\REQUIRE Story prompt $x$, configuration $c$, optional layout $L^{user}$, optional references $R^{user}$, optional template library $\mathcal{T}$
\ENSURE Generated manga $\hat{M}$
\STATE $Y \leftarrow A_{\mathrm{planner}}(x,c)$
\STATE Enforce configured page number and target panel counts in $Y$
\STATE Extract story sections $\mathcal{S}=\{s_k\}_{k=1}^{K}$
\FOR{each story section $s_k \in \mathcal{S}$}
    \STATE Load cached section memory $\mathcal{M}_k$ if available
    \IF{$\mathcal{M}_k$ is unavailable}
        \STATE Generate or assign scene, character, and key-object references
        \STATE Build $\mathcal{M}_k=(d_k,R_k^{scene},R_k^{char},R_k^{obj},\phi_k)$
    \ENDIF
\ENDFOR
\FOR{each page $i=1,\ldots,N$}
    \IF{user layout $L_i^{user}$ is provided}
        \STATE $L_i \leftarrow L_i^{user}$
    \ELSIF{a matched template exists in $\mathcal{T}$}
        \STATE $L_i \leftarrow \mathrm{RetrieveTemplate}(q_i,n_i,\mathcal{T})$
    \ELSE
        \STATE $L_i \leftarrow A_{\mathrm{layout}}(q_i,n_i,c)$
    \ENDIF
    \STATE $\tilde{L}_i \leftarrow \Pi(L_i)$
    \FOR{each panel $j=1,\ldots,n_i$}
        \STATE $k \leftarrow z(i,j)$
        \STATE $p_{i,j} \leftarrow A_{\mathrm{panel}}(y_{i,j},\tilde{L}_i,\mathcal{M}_k)$
        \STATE $R_{i,j} \leftarrow \mathrm{ComposeRef}(\mathcal{M}_k,R^{user})$
        \STATE $\hat{I}_{i,j} \leftarrow \mathcal{R}(p_{i,j},R_{i,j},w_{i,j},h_{i,j})$
    \ENDFOR
    \STATE $P_i^{img} \leftarrow \mathrm{ComposePage}(\{\hat{I}_{i,j}\},\tilde{L}_i)$
    \STATE $\hat{P}_i \leftarrow A_{\mathrm{text}}(P_i^{img},Y,\tilde{L}_i)$
\ENDFOR
\STATE $\hat{M} \leftarrow \mathrm{ComposeComic}(\{\hat{P}_i\}_{i=1}^{N})$
\RETURN $\hat{M}$
\end{algorithmic}
\end{algorithm}

\subsection{Details of Lettering and Readability Evaluation}
\label{app:lettering_readability}

\subsubsection{Bubble Placement Annotation}
\label{app:bubble_annotation}

We use human annotation to evaluate bubble placement because automatic methods cannot reliably determine whether a speech bubble occludes an important facial region in stylized manga images. 
Annotators evaluate each panel containing text bubbles. 
A panel is assigned a binary score:
\begin{equation}
s_p =
\begin{cases}
0, & \text{if any bubble occludes a character's face}, \\
1, & \text{otherwise}.
\end{cases}
\end{equation}
The final Bubble Placement Score is the average over all panels containing text bubbles:
\begin{equation}
\mathrm{BPS}
=
\frac{1}{|\mathcal{P}^{txt}|}
\sum_{p \in \mathcal{P}^{txt}} s_p.
\end{equation}
We only penalize face occlusion caused by text bubbles. 
Bubbles crossing panel boundaries or gutters are allowed as long as they do not damage the readability of the page or occlude character faces.

\subsubsection{Human Annotation for Bubble Placement}
\label{app:human_annotation}

We use human annotation only for evaluating bubble placement quality. The annotation task does not involve collecting personal information from annotators or evaluating human subjects. Annotators are asked to inspect generated manga panels that contain text bubbles and determine whether any bubble occludes a character's face.

We recruited $3$ annotators for this evaluation. Each panel containing text bubbles was evaluated by $3$ annotators, and the final Bubble Placement Score is computed by averaging scores over annotators and evaluated panels.

The annotation instruction given to annotators was:

\begin{quote}
Please inspect each manga panel containing one or more text bubbles. Assign a score of 1 if no text bubble occludes a character's face. Assign a score of 0 if any text bubble covers or substantially occludes a character's face. Only penalize face occlusion caused by text bubbles. Bubbles crossing panel boundaries or gutters should not be penalized unless they damage readability or occlude character faces.
\end{quote}

Annotators were informed about the purpose of the annotation task and gave consent to participate. They received no monetary compensation. The task only requires annotators to judge generated images and does not ask for any sensitive personal information.  
If multiple annotators evaluated the same panel, we report the mean score across annotators and samples. In cases of disagreement, we use the average score. The annotation results are used only for aggregate evaluation and are not used to train any model.

\subsubsection{LLM-based Manga Readability Evaluation}
\label{app:llm_readability}

We evaluate manga readability using a multimodal LLM judge. 
Instead of asking the judge for an overall preference score, we evaluate whether the generated manga can faithfully communicate the intended story. 
The evaluation is performed in two steps. 
First, the judge reads the generated manga pages and writes a concise summary of the perceived story. 
Second, the judge compares this generated summary with the ground-truth story and assigns a readability score from 1 to 5.

The scoring rubric is as follows:
\begin{itemize}
    \item \textbf{1}: The generated manga is largely unreadable or fails to communicate the ground-truth story.
    \item \textbf{2}: The manga conveys only a few isolated story elements, but the main plot is missing or incorrect.
    \item \textbf{3}: The manga conveys the rough story direction, but important events, character actions, or causal relations are missing.
    \item \textbf{4}: The manga communicates most of the ground-truth story with minor omissions or ambiguities.
    \item \textbf{5}: The manga clearly and faithfully communicates the ground-truth story.
\end{itemize}

We use the following prompt template for the multimodal LLM judge:

\begin{lstlisting}[style=promptstyle]
You are evaluating the readability of a generated manga.

First, carefully read the provided manga pages and summarize the story conveyed by the manga.
Then compare your summary with the ground-truth story.
Assign a score from 1 to 5 according to how faithfully the manga communicates the ground-truth story.

Use a strict story-overlap criterion, not a generous impression score.
Estimate how much of the ground-truth story is communicated by the manga as an overlap percentage from 0 to 100.
Base this overlap on the ground-truth story elements that a reader can actually infer from the pages:
- main characters and their identities/roles;
- key settings and setting transitions;
- important events and actions;
- causal relations between events;
- temporal/order relations;
- ending state or resolution.

Important judging rules:
- Ambiguous, unreadable, or visually unclear content counts as missing.
- Incorrectly depicted events count as missing, even if the image is visually plausible.
- Extra unrelated events do not compensate for missing ground-truth events.
- Do not reward image quality, style, or aesthetics unless they directly improve story communication.
- If you are uncertain between two scores, choose the lower score.
- The score must exactly follow the overlap thresholds below.

Scoring rubric:
1: overlap < 40%. The manga is largely unreadable or fails to communicate the ground-truth story.
2: 40% <= overlap < 60%. The manga conveys some isolated or partial story elements, but the main plot is substantially missing or incorrect.
3: 60% <= overlap < 80%. The manga conveys the rough story direction, but important events, character actions, settings, or causal/order relations are missing.
4: 80% <= overlap < 90%. The manga communicates most of the ground-truth story, with only moderate omissions or ambiguities.
5: overlap >= 90%. The manga clearly and faithfully communicates almost all of the ground-truth story; only very minor details may be missing.

Ground-truth story:
{ground_truth_story}

Return your answer as strict JSON with this schema:
{
  "summary": "...",
  "overlap_percent": integer from 0 to 100,
  "matched_elements": ["..."],
  "missing_or_wrong_elements": ["..."],
  "score": 1,
  "reason": "..."
}
\end{lstlisting}

\subsection{Additional Experiment Analysis}

\begin{table*}[t]
\centering
\scriptsize
\setlength{\tabcolsep}{3.5pt}
\caption{
Quantitative results on ViStoryBench. 
We compare direct page generation baselines with MangaFlow using different rendering backbones. 
CSD denotes style similarity, CIDS denotes character identity similarity, and PA denotes prompt alignment. 
For PA, Scene, Shot, CI, and IA denote scene alignment, camera/shot alignment, character-action alignment, and individual-action alignment, respectively. 
CM denotes OCCM, Inc denotes inception score, Aes denotes aesthetic score, and CP denotes copy-paste score. 
}
\label{tab:main_vistorybench}
\begin{tabular*}{\textwidth}{@{\extracolsep{\fill}}llccccccccccccc@{}}
\toprule
\multirow{2}{*}{Method} 
& \multirow{2}{*}{Model}
& \multicolumn{2}{c}{CSD$\uparrow$}
& \multicolumn{2}{c}{CIDS$\uparrow$}
& \multicolumn{5}{c}{PA$\uparrow$}
& \multirow{2}{*}{CM$\uparrow$}
& \multirow{2}{*}{Inc$\uparrow$}
& \multirow{2}{*}{Aes$\uparrow$}
& \multirow{2}{*}{CP$\downarrow$} \\
\cmidrule(lr){3-4}
\cmidrule(lr){5-6}
\cmidrule(lr){7-11}
& 
& Cross & Self 
& Cross & Self 
& Scene & Shot & CI & IA & Avg. 
& & & & \\
\midrule
Direct 
& Gemini 
& 0.373 
& 0.702 
& 0.454 
& 0.562 
& {3.253} 
& 2.762 
& 2.846 
& 1.971 
& 2.708 
& 51.60 
& 8.14 
& 4.52 
& 0.493 \\

Direct 
& FLUX.2 
& 0.229 
& \textbf{0.750} 
& 0.358 
& 0.591 
& 2.280 
& 2.620 
& 1.885 
& 1.363 
& 2.037 
& 39.05 
& 2.24 
& 4.56 
& 0.493 \\

MangaFlow 
& Gemini 
& \textbf{0.383} 
& 0.622 
& \textbf{0.589} 
& \textbf{0.643} 
& 3.061 
& 2.912 
& 3.163 
& 2.468 
& 2.901 
& \textbf{73.93} 
& \textbf{12.26} 
& 5.54 
& 0.506 \\

MangaFlow 
& FLUX.2 
& 0.328 
& 0.668 
& 0.447 
& 0.619 
& \textbf{3.861} 
& \textbf{3.008} 
& \textbf{3.638} 
& \textbf{2.688} 
& \textbf{3.299} 
& 66.87 
& 10.61 
& \textbf{5.92} 
& \textbf{0.483} \\
\bottomrule
\end{tabular*}
\end{table*}

\begin{table*}[h]
\centering
\scriptsize
\setlength{\tabcolsep}{3.5pt}
\caption{
Ablation study on Story Section Memory on ViStoryBench. 
We compare the full MangaFlow-FLUX pipeline with the variant without Story Section Memory. 
CSD denotes style similarity, CIDS denotes character identity similarity, and PA denotes prompt alignment. 
For PA, Scene, Shot, CI, and IA denote scene alignment, camera/shot alignment, character-action alignment, and individual-action alignment, respectively. 
CM denotes OCCM, Inc denotes inception score, Aes denotes aesthetic score, and CP denotes copy-paste score.
}
\label{tab:fullablation_story_section_memory}
\begin{tabular*}{\textwidth}{@{\extracolsep{\fill}}lccccccccccccc@{}}
\toprule
\multirow{2}{*}{Variant} 
& \multicolumn{2}{c}{CSD$\uparrow$}
& \multicolumn{2}{c}{CIDS$\uparrow$}
& \multicolumn{5}{c}{PA$\uparrow$}
& \multirow{2}{*}{CM$\uparrow$}
& \multirow{2}{*}{Inc$\uparrow$}
& \multirow{2}{*}{Aes$\uparrow$}
& \multirow{2}{*}{CP$\downarrow$} \\
\cmidrule(lr){2-3}
\cmidrule(lr){4-5}
\cmidrule(lr){6-10}
& Cross & Self 
& Cross & Self 
& Scene & Shot & CI & IA & Avg. 
& & & & \\
\midrule
w/o Sec. Mem. 
& {0.327} 
& 0.547 
& {0.442} 
& 0.582 
& {3.836} 
& \textbf{3.176} 
& {3.439} 
& {2.525} 
& {3.244} 
& \textbf{69.41} 
& \textbf{12.91} 
& 5.83 
& {0.503} \\

Full MangaFlow-FLUX 
& \textbf{0.328} 
& \textbf{0.668} 
& \textbf{0.447} 
& \textbf{0.619} 
& \textbf{3.861} 
& 3.008
& \textbf{3.638} 
& \textbf{2.688} 
& \textbf{3.299} 
& 66.87 
& 10.61 
& \textbf{5.92} 
& \textbf{0.483} \\
\bottomrule
\end{tabular*}
\end{table*}

\subsubsection{Story Section Memory Ablation Visualization}

\label{app:section_memory_vis}

\begin{figure*}[h]
    \centering

    \begin{minipage}[t]{0.155\linewidth}
        \centering
        \includegraphics[
            width=\linewidth,
            page=1,
            trim={0 0 0 0},
            clip
        ]{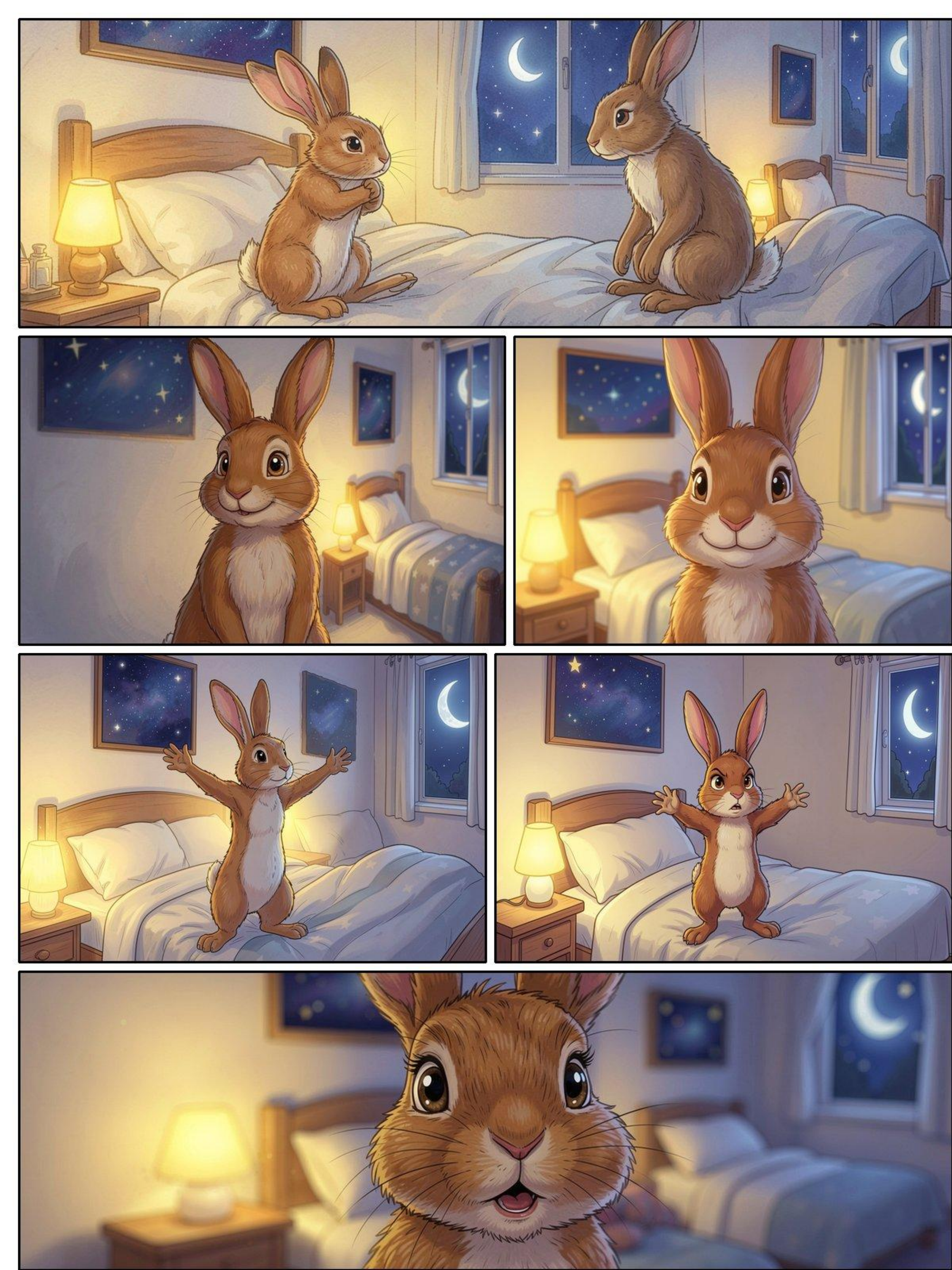}
        \vspace{2pt}
        
        {\scriptsize (a) Example 1 ours}
    \end{minipage}
    \hfill
    \begin{minipage}[t]{0.155\linewidth}
        \centering
        \includegraphics[
            width=\linewidth,
            page=2,
            trim={0 0 0 0},
            clip
        ]{figures/memory/merged_stories.pdf}
        \vspace{2pt}
        
        {\scriptsize (b) Example 1 w/o memory}
    \end{minipage}
    \hfill
    \begin{minipage}[t]{0.155\linewidth}
        \centering
        \includegraphics[
            width=\linewidth,
            page=3,
            trim={0 0 0 0},
            clip
        ]{figures/memory/merged_stories.pdf}
        \vspace{2pt}
        
        {\scriptsize (c) Example 2 ours}
    \end{minipage}
    \hfill
    \begin{minipage}[t]{0.155\linewidth}
        \centering
        \includegraphics[
            width=\linewidth,
            page=4,
            trim={0 0 0 0},
            clip
        ]{figures/memory/merged_stories.pdf}
        \vspace{2pt}
        
        {\scriptsize (d) Example 2 w/o memory}
    \end{minipage}
    \hfill
    \begin{minipage}[t]{0.155\linewidth}
        \centering
        \includegraphics[
            width=\linewidth,
            page=6,
            trim={0 0 0 0},
            clip
        ]{figures/memory/merged_stories.pdf}
        \vspace{2pt}
        
        {\scriptsize (e) Example 3 ours}
    \end{minipage}
    \hfill
    \begin{minipage}[t]{0.155\linewidth}
        \centering
        \includegraphics[
            width=\linewidth,
            page=5,
            trim={0 0 0 0},
            clip
        ]{figures/memory/merged_stories.pdf}
        \vspace{2pt}
        
        {\scriptsize (f) Example 3 w/o memory}
    \end{minipage}

    \caption{
    Visualization of Story Section Memory. 
    The pages show that MangaFlow can maintain consistent recurring characters, scenes, and key objects across panels and pages by reusing section-level textual and visual references.
    }
    \label{fig:section_memory_vis}
\end{figure*}

We provide qualitative examples to illustrate the effect of Story Section Memory. 
The visualization shows that reusing section-level character and scene references helps MangaFlow maintain more stable visual identities across panels and pages. 
As shown in Figure~\ref{fig:section_memory_vis}, recurring characters, background scenes, and important objects remain more consistent throughout the generated manga.

\subsubsection{Renderer Visualization}
\label{app:renderer_vis}

\begin{figure*}[h!]
    \centering

    \begin{minipage}[t]{0.24\linewidth}
        \centering
        \includegraphics[
            width=\linewidth,
            height=1.42\linewidth,
            page=1,
            trim={0 0 0 0},
            clip
        ]{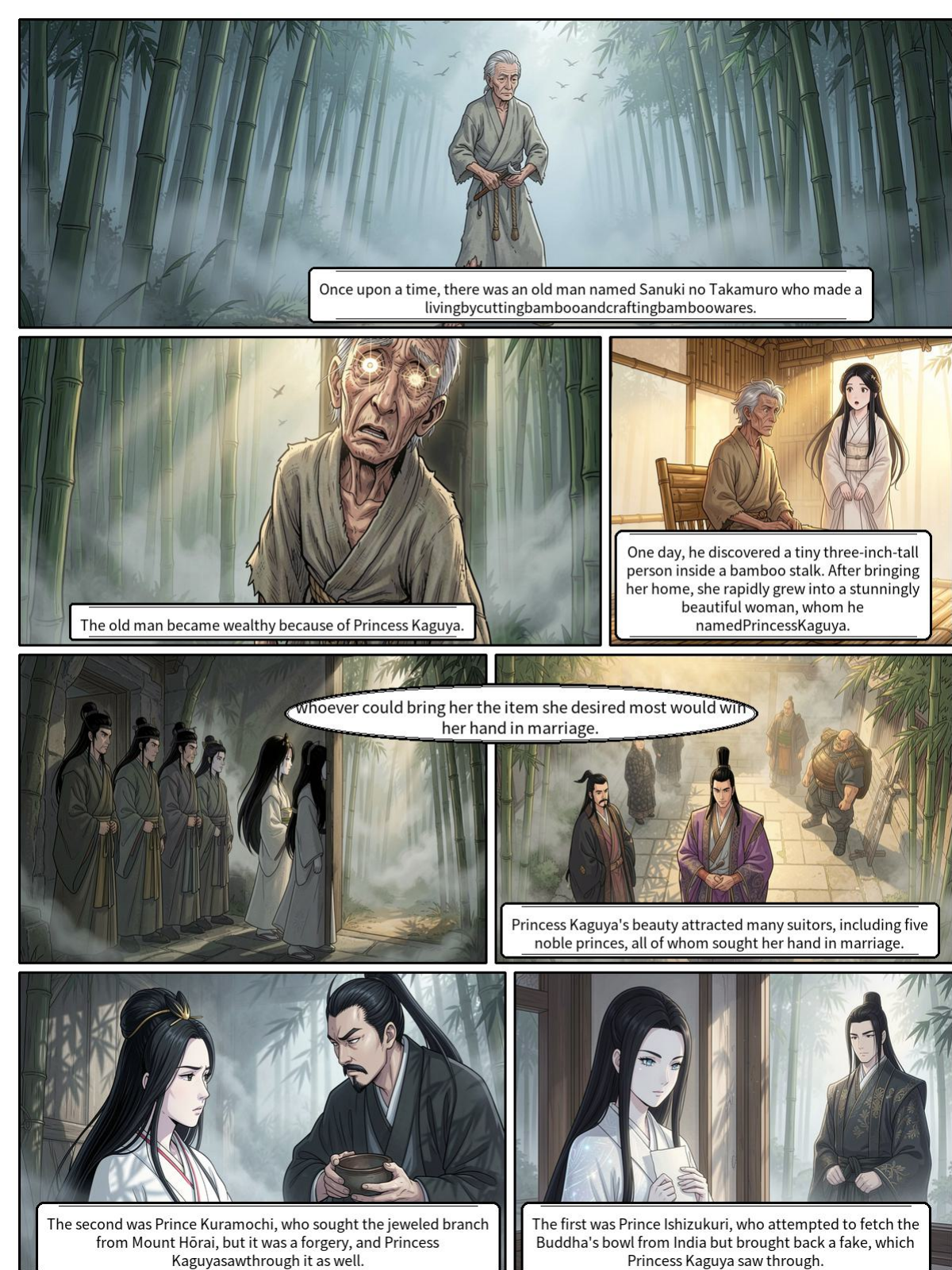}
        
        \vspace{2pt}
        {\scriptsize FLUX.2, Page 1}
    \end{minipage}
    \hfill
    \begin{minipage}[t]{0.24\linewidth}
        \centering
        \includegraphics[
            width=\linewidth,
            height=1.42\linewidth,
            page=2,
            trim={0 0 0 0},
            clip
        ]{figures/render/flux2.pdf}
        
        \vspace{2pt}
        {\scriptsize FLUX.2, Page 2}
    \end{minipage}
    \hfill
    \begin{minipage}[t]{0.24\linewidth}
        \centering
        \includegraphics[
            width=\linewidth,
            height=1.42\linewidth,
            page=1,
            trim={0 0 0 0},
            clip
        ]{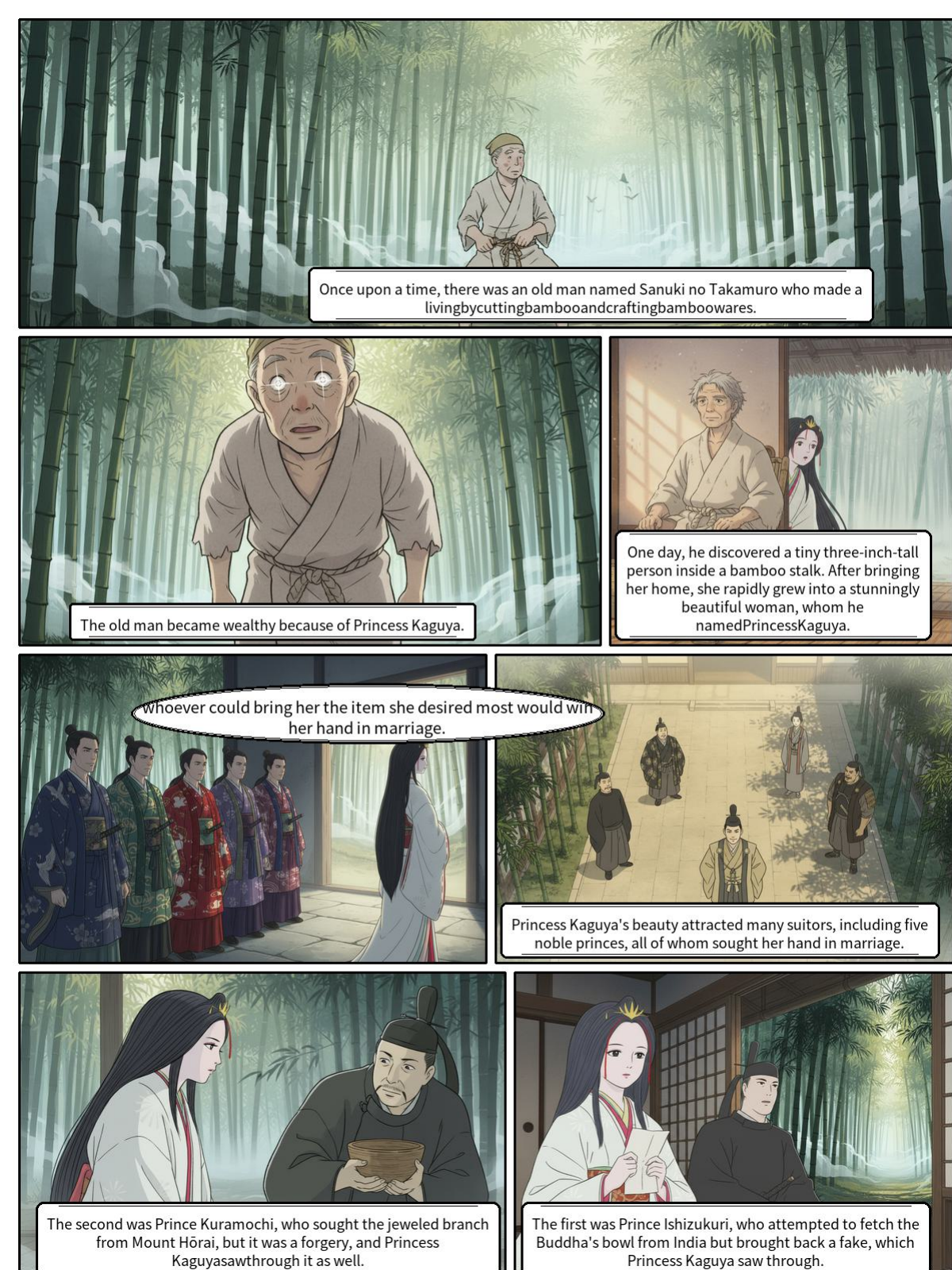}
        
        \vspace{2pt}
        {\scriptsize Gemini, Page 1}
    \end{minipage}
    \hfill
    \begin{minipage}[t]{0.24\linewidth}
        \centering
        \includegraphics[
            width=\linewidth,
            height=1.42\linewidth,
            page=2,
            trim={0 0 0 0},
            clip
        ]{figures/render/nano.pdf}
        
        \vspace{2pt}
        {\scriptsize Gemini, Page 2}
    \end{minipage}

    \caption{
    Renderer comparison within MangaFlow. 
    The same story plan, story-section memory, layout, and page composition are used, while the renderer is changed between FLUX.2 and Gemini 2.5 flash image. 
    This comparison visualizes how different image generation backbones affect the final manga appearance under the same structured pipeline.
    }
    \label{fig:renderer_vis}
\end{figure*}

We provide additional qualitative examples as shown in Fig.\ref{fig:renderer_vis} to compare different rendering backbones within the same MangaFlow pipeline. 
The story plan, story-section memory, layout, and page composition are kept unchanged, while only the panel renderer is changed. 
This visualization helps isolate the effect of the rendering backbone on visual appearance, consistency, and panel-level image quality.

\subsubsection{Text-to-Manga Generation from a Simple Prompt}
\label{app:text2manga_demo}

To further demonstrate the end-to-end generation capability of MangaFlow, we provide an example of text-to-manga generation from a simple story prompt. 
Unlike settings that require detailed storyboards, predefined layouts, or manually provided visual references, this example only uses a short textual prompt as input. 
MangaFlow automatically decomposes the prompt into a structured story plan, constructs story sections, generates page layouts, renders individual panels, composes manga pages, and places dialogue text.

Figure~\ref{fig:text2manga_demo} shows the generated four-page manga. 
The result demonstrates that MangaFlow can function as a complete text-to-manga system: starting from a natural-language prompt, it produces multi-page manga with coherent story progression, explicit panel structure, page composition, and lettering. 
This ability complements the controllable setting discussed in the main paper, where users may additionally provide story JSON files, reference images, or target layouts for more precise control.

The story prompt of Fig.\ref{fig:text2manga_demo} is shown below:

\begin{lstlisting}[style=promptstyle]
After school, a boy named Ren and a girl named Mika enter a quiet classroom. As soon as they step inside, a mysterious light fills the room, and the classroom transforms into a magical gateway. They are suddenly transported to China, where they walk through crowded streets, ancient buildings, red lanterns, and food stalls. Before they can react, the scene changes to the United States, with tall skyscrapers, wide avenues, and bright city lights. Then they travel to Japan, surrounded by cherry blossoms, traditional shrines, and modern city streets. Finally, they appear in France, standing near elegant streets, art museums, and the Eiffel Tower.
Suddenly, the sky and scenery begin to twist and fade. The boy and girl are pulled back through the glowing portal and return to the classroom. They wake up at their desks with a shock, breathing quickly and looking around in confusion. The classroom is quiet again, but both of them remember the same impossible journey. They stare at each other, unsure whether it was only a dream or a real adventure.
\end{lstlisting}
\subsubsection{Statistical Significance of the Paired Story-Level Comparison}
\label{sec:stats}

We further conduct paired story-level significance tests, treating each story as
the sampling unit. We report $95\%$ confidence intervals from $1{,}000$
bootstrap resamples and use two-sided Wilcoxon signed-rank tests with Holm
correction; paired $t$-tests are used as a sensitivity
analysis. 
As shown in Table~\ref{tab:stats}, the representative improvements consistently
exclude zero and remain significant after multiple-comparison correction.  

\begin{table*}[t!h]
\centering
\small
\setlength{\tabcolsep}{4pt}
\caption{Paired story-level improvement of MangaFlow over the Direct baseline,
one representative metric per evaluation dimension. Confidence intervals are
estimated from $1{,}000$ bootstrap resamples, and $p$-values are from
two-sided Wilcoxon signed-rank tests with Holm correction over all $34$
comparisons.}
\label{tab:stats}
\begin{tabular}{@{}llrr@{}}
\toprule
\textbf{Renderer} & \textbf{Dimension (metric)} &
\textbf{$\Delta$ (MangaFlow $-$ Direct) [95\% CI]} & \textbf{Holm $p$} \\
\midrule
FLUX   & Character consistency (CIDS cross) & $+0.0893 [0.0752, 0.1031]$ & $1.06{\times}10^{-12}$ \\
FLUX   & Layout adherence (IoU)             & $+0.5820$ $[0.5580, 0.6046]$ & $2.59{\times}10^{-13}$ \\
FLUX   & Character--action alignment        & $+1.7528 [1.6319, 1.8712]$ & $3.32{\times}10^{-13}$ \\
FLUX   & Readability                        & $+2.0380$ $[1.8731, 2.2025]$ & $1.23{\times}10^{-13}$ \\
\midrule
Gemini & Character consistency (CIDS cross) & $+0.1391$ $[0.1037, 0.1779]$ & $8.54{\times}10^{-8}$ \\
Gemini & Layout adherence (IoU)             & $+0.5684$ $[0.5368, 0.5993]$ & $1.67{\times}10^{-10}$ \\
Gemini & Character--action alignment        & $+0.3121$ $[0.1978, 0.4324]$ & $6.53{\times}10^{-5}$ \\
Gemini & Readability                        & $+0.9355$ $[0.7258, 1.1613]$ & $3.73{\times}10^{-7}$ \\
\bottomrule
\end{tabular}
\end{table*}

\subsubsection{Human--LLM Calibration Study for Readability}
\label{sec:human}

To validate the GPT-4.1 readability metric, we conduct a blind human study on
$50$ samples from $25$ paired MangaFlow-Gemini and Direct-Gemini stories.
Three annotators independently score each sample using the same $1$--$5$
readability rubric as GPT-4.1.

As shown in Table~\ref{tab:human-calib}, human annotators are systematically
stricter than GPT-4.1 on the absolute scale, with a mean human-minus-LLM
difference of $-1.25$. However, the relative method comparison is highly
consistent: humans prefer MangaFlow in $24/25$ stories, while GPT-4.1 never
prefers Direct. The human--LLM no-conflict agreement is $96.0\%$. We therefore
use GPT-4.1 readability primarily as a relative comparison between methods,
rather than as a calibrated absolute readability score.

\begin{table*}[!th]
\centering
\small
\caption{Human--LLM calibration study for readability
($50$ samples from $25$ paired stories, three annotators). Human ratings are
stricter on the absolute scale but strongly agree with GPT-4.1 on the relative
method comparison.}
\label{tab:human-calib}
\begin{tabular}{@{}lrrr@{}}
\toprule
\multicolumn{4}{@{}l}{\emph{(a) Absolute score calibration}} \\
\midrule
\textbf{Subset} & \textbf{Human} & \textbf{GPT-4.1} & \textbf{Human $-$ LLM} \\
All samples ($n{=}50$)      & $3.11$ & $4.36$ & $-1.25$ \\
MangaFlow-Gemini ($n{=}25$) & $3.73$ & $4.88$ & $-1.15$ \\
Direct-Gemini ($n{=}25$)    & $2.49$ & $3.84$ & $-1.35$ \\
\midrule
\multicolumn{4}{@{}l}{\emph{(b) Agreement}} \\
\midrule
Human--LLM rank correlation (Spearman)       & \multicolumn{3}{r}{$0.562$} \\
Inter-annotator within-one-point agreement   & \multicolumn{3}{r}{$84.0\%$} \\
\midrule
\multicolumn{4}{@{}l}{\emph{(c) Paired preference over $25$ stories}} \\
\midrule
Averaged human judgement (MangaFlow / Direct / tie) & \multicolumn{3}{r}{$24$ / $1$ / $0$} \\
GPT-4.1 judgement (MangaFlow / Direct / tie)         & \multicolumn{3}{r}{$17$ / $0$ / $8$} \\
Human--LLM no-conflict agreement                     & \multicolumn{3}{r}{$24/25 = 96.0\%$} \\
\bottomrule
\end{tabular}
\end{table*}

\subsubsection{Visualization of Self-Reflection}
\label{app:self_reflection_vis}

\begin{figure*}[h]
    \centering
    \includegraphics[
        width=0.95\textwidth,
        trim={0 300pt 0 300pt},
        clip
    ]{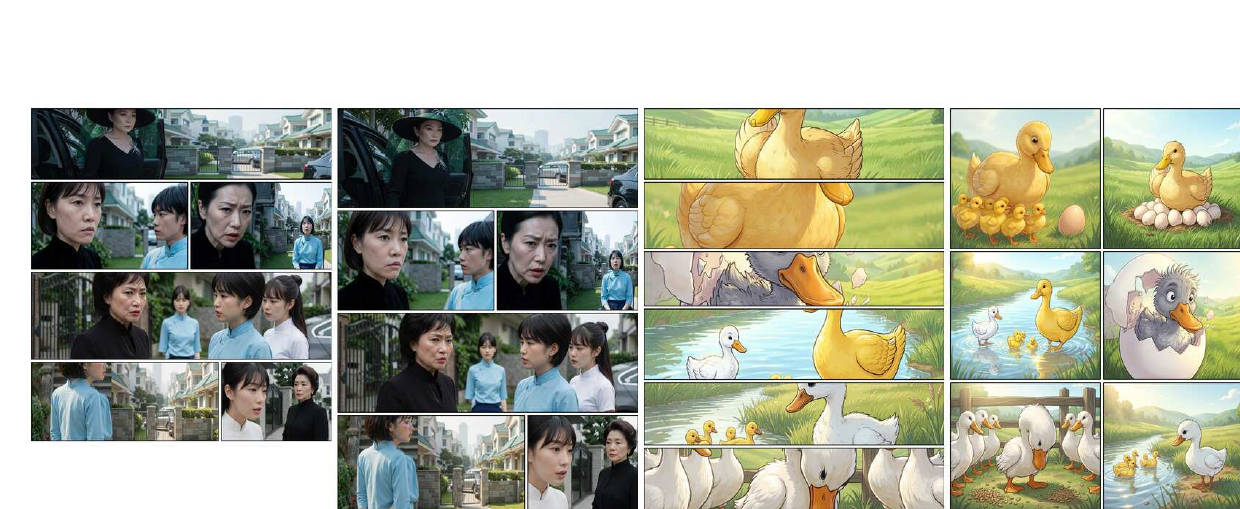}

    \caption{
    Visualization of layout self-reflection in MangaFlow. 
    Each row shows one example, where the left page is generated using the original layout and the right page is generated using the self-refined layout.
    The reflected layouts better balance panel allocation and page-space usage while preserving the original story content.
    }
    \label{fig:self_reflection_vis}
\end{figure*}

To better understand the effect of self-reflection in MangaFlow, we provide qualitative visualizations of the layout changes introduced by the Layout Agent. 
Each example in Fig. \ref{fig:self_reflection_vis} contains two pages: the left page shows the original layout before self-reflection, while the right page shows the reflected layout after self-reflection. 
The visualization demonstrates how self-reflection helps reorganize panel allocation, reduce layout imbalance, and improve page-space usage without changing the underlying story content.

\subsection{Discussion and Limitations}
\label{sec:discussion_limitations}

MangaFlow demonstrates that controllable manga generation benefits from explicit intermediate structures, but several limitations remain. 
First, the final visual quality still depends on the underlying image generation backbone, so failures in character rendering, fine-grained actions, or local visual details may affect the generated panels. 
Second, although MangaFlow introduces explicit layout control and lettering, speaker-aware bubble placement remains challenging in complex panels, especially when stylized manga faces are difficult to localize. 
Third, MangaGen-MetaBench is designed as a meta-benchmark built on existing story visualization data, rather than a fully annotated manga dataset with human-designed layouts, bubble positions, and speaker annotations.

These limitations suggest several future directions. 
More powerful reference-conditioned renderers could further improve character and scene consistency. 
Manga-specific perception models may help improve face localization, speaker grounding, and bubble placement. 
Richer manga annotations could also support more comprehensive evaluation of layout design, reading flow, and lettering quality. 
Despite these limitations, MangaFlow provides an initial step toward a controllable agentic paradigm for manga generation.

\end{document}